\def\namedcopy{}
\documentclass{article}

\PassOptionsToPackage{table}{xcolor}

\usepackage[main, preprint]{neurips_2026}

\usepackage[utf8]{inputenc}
\usepackage[T1]{fontenc}
\usepackage{amsmath,amssymb,amsfonts}
\usepackage{graphicx}
\graphicspath{{./}{figures/}{../figures/}{../../figures/}}
\usepackage{booktabs}
\usepackage{xspace}
\usepackage{enumitem}
\usepackage[hidelinks]{hyperref}
\usepackage{xurl}
\usepackage{microtype}
\usepackage{tikz}
\usetikzlibrary{fit, positioning, arrows.meta, calc}
\usepackage[table]{xcolor}
\definecolor{rankblue}{HTML}{4477AA}
\definecolor{flatorange}{HTML}{CC6677}
\definecolor{bilgreen}{HTML}{117733}
\definecolor{flatred}{HTML}{882255}
\definecolor{atom}{HTML}{1F77B4}
\definecolor{comparison}{HTML}{2CA02C}
\definecolor{baseline}{HTML}{9467BD}
\definecolor{control}{HTML}{17BECF}
\definecolor{negative}{HTML}{D62728}
\definecolor{auxiliary}{HTML}{FF7F0E}

\usepackage{multirow}
\usepackage{placeins}
\usepackage{float}
\usepackage{algorithm}
\usepackage{algpseudocode}
\usepackage[font=footnotesize,skip=4pt]{caption}

\makeatletter
\renewcommand\section{\@startsection {section}{1}{\z@}%
                     {-1.5ex \@plus -0.4ex \@minus -0.2ex}%
                     {1.0ex \@plus 0.2ex \@minus 0.1ex}%
                     {\large\bf\raggedright}}
\renewcommand\subsection{\@startsection{subsection}{2}{\z@}%
                        {-1.2ex \@plus -0.4ex \@minus -0.2ex}%
                        {0.5ex \@plus 0.1ex}%
                        {\normalsize\bf\raggedright}}
\renewcommand\paragraph{\@startsection{paragraph}{4}{\z@}%
                       {1.0ex \@plus 0.4ex \@minus 0.2ex}%
                       {-1em}%
                       {\normalsize\bf}}
\makeatother

\ifdefined\namedcopy
  \newcommand{\repourl}{https://github.com/JackYoung27/writesae}
  \newcommand{\hfhandle}{jackyoung27}
\else
  \newcommand{\repourl}{https://anonymous.4open.science/r/WriteSAE-6158/}
  \newcommand{\hfhandle}{anon-writesae}
\fi

\makeatletter
\newcommand{\repolink}{%
  \edef\msae@tmp{\repourl}%
  \ifx\msae@tmp\@empty\textit{[anonymized supplementary]}%
  \else\url{\repourl}\fi\xspace}
\newcommand{\hfrepolink}[1]{%
  \edef\msae@tmp{\hfhandle}%
  \ifx\msae@tmp\@empty\textit{[anonymized supplementary]}%
  \else\href{https://huggingface.co/\hfhandle/writesae-ckpts/tree/main/#1}{\texttt{\hfhandle/writesae-ckpts/#1}}\fi\xspace}
\newcommand{\hfreporootlink}{%
  \edef\msae@tmp{\hfhandle}%
  \ifx\msae@tmp\@empty\textit{[anonymized supplementary]}%
  \else\href{https://huggingface.co/\hfhandle/writesae-ckpts}{\texttt{\hfhandle/writesae-ckpts}}\fi\xspace}
\makeatother

\newcommand{\flatsae}{FlatSAE\xspace}
\newcommand{\writesae}{WriteSAE\xspace}
\newcommand{\ranksae}{MatrixSAE\xspace}
\newcommand{\bilinearsae}{BilinearSAE\xspace}
\newcommand{\bilinsae}{\bilinearsae}
\newcommand{\gdn}{Gated DeltaNet\xspace}

\newcommand{\state}{S_t}

\newcommand{\nf}{n_f}
\newcommand{\Sref}[1]{Section~\ref{#1}}

\title{WriteSAE: Sparse Autoencoders for Recurrent State}

\ifdefined\namedcopy
  \author{Jack Young \\
Indiana University \\
\href{mailto:youngjh@iu.edu}{youngjh@iu.edu}
}
\else
  \author{Anonymous submission}
\fi

\begin{document}

\maketitle

\begin{abstract}
We introduce \writesae, a sparse autoencoder for the matrix updates written into
recurrent language-model state. In Gated DeltaNet, Mamba-2, and RWKV-7, each
token writes a matrix-shaped update to a recurrent cache; a residual-stream SAE
has vector-shaped atoms and cannot replace that update directly. \writesae
learns rank-1 matrix atoms with the same shape as the model's own write. This
lets us test a direct replacement: at positions where the SAE activates an atom,
we remove the model's write, insert the atom scaled by its SAE activation, and
continue the forward pass. The atom gives a closer final token distribution than
deleting the write on $\mathbf{92.4\%}$ of evaluated positions; averaged per
atom, the rate is $89.8\%$. For Gated DeltaNet, a formula using the forget gate,
read query, and output embedding predicts the resulting logit change with
$R^2{=}0.98$. The same replacement test transfers to Mamba-2-370M at $88.1\%$.
In generation, the formula chooses a write direction; writing it into three
consecutive cache positions at $3\!\times\!$ the norm of the model's write makes
tokens initially ranked $100$--$1000$ by the unmodified model appear in
$\mathbf{100\%}$ of continuations, up from $33.3\%$. To our knowledge this is
the first cache-level steering intervention reported in a state-space or hybrid
recurrent layer.
\end{abstract}

\section{Introduction}
\label{sec:intro}

State-space and hybrid recurrent models (Mamba-2, RWKV-7, Gated DeltaNet, Qwen3.5) write memory into a matrix cache rather than only into the residual stream. In the GDN recurrence of \citet{yang2024gdn}, each token adds one rank-1 matrix $\mathbf{k}_t\mathbf{v}_t^\top$ to a $d_k \times d_v$ state; later tokens read that state with a query $\mathbf{q}_{t'}$. Across a long context, many updates share the same state, and superposition theory predicts overlap among the features carrying them \citep{elhage2022toymodels, scherlis2022polysemanticity}.

\begin{figure}[!b]
  \centering
  \scalebox{0.85}{\input{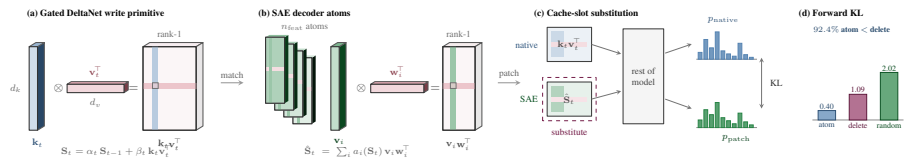}}
  \caption{\textbf{\writesae atoms can replace native \gdn writes.} At a primary GDN layer-head, atoms beat deleting the write on $92.4\%$ of evaluated token positions; panels show the model write $\mathbf{k}_t\mathbf{v}_t^\top$, the learned atom $\mathbf{v}_i\mathbf{w}_i^\top$, the replacement test, and the KL controls.}
  \label{fig:method_schematic}
\end{figure}

Residual SAEs \citep{bricken2023monosemanticity, cunningham2024saes, templeton2024scaling, gao2024scaling} and Mamba/RWKV extensions \citep{wang2024universality, paulo2024rnntransfer, hossain2025mambamemory, sunkumohan2026asb} analyze vectors produced by a layer. Here the object is the matrix state that later tokens read. A standard SAE can be trained on $\mathrm{vec}(\state)$, but its decoder atoms are $d_kd_v$-vectors. Replacing one model update requires an outer product because the next layer reads the state with a query. Fast-weight work already casts the per-token update as rank-1 \citep{schmidhuber1992fwp, ba2016fastweights, schlag2021fwp}; \writesae uses the same shape for the dictionary.

\writesae decoder atoms are rank-1 matrices $\mathbf{v}_i \mathbf{w}_i^\top$ shaped like GDN's $\mathbf{k}_t \mathbf{v}_t^\top$, so a single atom can replace one cache update while preserving the shape read downstream (Fig.~\ref{fig:method_schematic}). A \emph{firing} is a token position where the SAE assigns a nonzero coefficient to an atom. The atoms split into two behaviors: some align tightly with a single native write at the positions where they fire; others spread their direction across many writes. \Sref{sec:classes} names these populations and reports the Gaussian-mixture fit.

\paragraph{Contributions.}
\begin{enumerate}[topsep=2pt,itemsep=1pt,leftmargin=*,label=(\arabic*)]
\item A replacement test in which a learned rank-1 atom replaces the native \gdn write; atoms beat deleting the write on $92.4\%$ of evaluated positions (\Sref{sec:mechvalid}).
\item A formula for logit change, using the gate, read query, and output embedding, that predicts measured effects at median $R^2{=}0.98$ (\Sref{sec:falsification}).
\item Cache erasure and generation probes showing targeted logit and continuation changes where the formula is accurate (\Sref{sec:steering}).
\item The same replacement test on Mamba-2-370M, where atoms beat deleting the write on $88.08\%$ of evaluated positions (\Sref{sec:archscope}).
\end{enumerate}

The replacement test runs at a single GDN layer-head: replace the native write with one learned atom scaled by its SAE coefficient, compare against deleting the write and against a random atom with the same coefficient, and measure KL divergence at the final output distribution. Atoms beat deleting the write on $92.4\%$ of evaluated positions, and the per-atom average is $89.8\%$ (\Sref{sec:mechvalid}). We also derive a formula for how one rank-1 perturbation changes a candidate token's logit; it tracks measured effects at median $R^2{=}0.98$ (\Sref{sec:falsification}).

The same formula supplies directions for three cache interventions (\Sref{sec:steering}). Erasing a single atom's contribution on its firing positions drops the promoted token's log probability by $0.116$ nats. Single-position edits have the predicted sign on $84.6\%$ of tested atom-token-context triples. During generation, writing the direction into three cache positions at $3{\times}$ makes tokens initially ranked $100$--$1000$ by the unmodified model appear in $\mathbf{100\%}$ of continuations, up from $33.3\%$, with $+1.27$ nats of first-step support.

Transfer depends on the model's write rule. The same replacement test reaches $88.08\%$ on Mamba-2-370M (\Sref{sec:archscope}). Across the tested recurrent families, the median cosine between an atom and the nearest native write at its firing positions is highest in GDN ($0.262$), then RWKV-7 ($0.180$), then Mamba-2 ($0.0575$). Four reported failures define the current scope: the larger Qwen model, the Mamba-2 closed-form logit shift, rank-2 atoms, and Mamba-2 generation edits. Code and checkpoints are at \repolink.

\section{Method}
\label{sec:method}

For \gdn, suppose we add a small rank-1 matrix $\varepsilon\,\mathbf{v}_i\mathbf{w}_i^\top$ to the cached state at position $t_0$. We predict the resulting change in the logit of token $\mathrm{tok}$ at a later position $t$ by
\begin{equation}
  \label{eq:closed_form}
  \Delta\ell_{\mathrm{tok}}(c, i, t)
    \approx G_{t_0 \to t}(c)\,
      \langle \mathbf{w}_i, \mathbf{q}_t(c) \rangle\,
      \langle \mathbf{v}_i, W_U[\mathrm{tok}] \rangle .
\end{equation}
Here $\Delta\ell_{\mathrm{tok}}(c,i,t)$ is the logit change in context $c$. Every quantity on the right is observable from a single forward pass. The scalar $G_{t_0\to t}(c)=\prod_{s=t_0+1}^{t}\alpha_s(c)$ is the product of the model's forget gates, $\mathbf{q}_t$ is the query that reads the cache, and $W_U[\mathrm{tok}]$ is the output-embedding row for the token. \Sref{sec:mechvalid} compares this formula with measured logit changes and obtains population $R^2{=}0.98$.

\paragraph{Where the expression comes from.}
\gdn writes one rank-$1$ outer $\mathbf{k}_t \mathbf{v}_t^\top$ into the matrix state $\state$ per token. The host recurrence is the gated delta rule of \citet{yang2024gdn},
\begin{equation}
  \label{eq:gdn_update}
  S_t = \alpha_t\bigl(I - \beta_t\,\mathbf{k}_t\mathbf{k}_t^\top\bigr) S_{t-1} + \beta_t \, \mathbf{k}_t \mathbf{v}_t^\top .
\end{equation}
Subtracting perturbed and native trajectories cancels the additive write at every later step, leaving
\begin{equation}
  \label{eq:state_perturb}
  \delta S_{s+1} = \alpha_{s+1}(c)\bigl(I - \beta_{s+1}(c)\,\mathbf{k}_{s+1}(c)\mathbf{k}_{s+1}(c)^\top\bigr)\delta S_s,\qquad
  \delta S_{t_0} = \varepsilon\,\mathbf{v}_i\mathbf{w}_i^\top ,
\end{equation}
so later steps only multiply the perturbation by matrices the model already computes. When $\mathbf{w}_i$ is nearly orthogonal to later keys, the extra term from $\mathbf{k}_{s+1}\mathbf{k}_{s+1}^\top$ is small; empirically this is the regime where the $R^2$ fit is high. Reading $\delta S_t$ with the query $\mathbf{q}_t$ and projecting to logits gives the two inner products in Eq.~\eqref{eq:closed_form}. App.~\ref{app:closed_form} gives the full derivation.

\paragraph{Atom rank matches host write rank.}
A decoder atom with the same rank as the model's update can replace one native rank-1 write while preserving the matrix shape. \writesae trains a TopK SAE whose decoder atoms factor as $\mathbf{v}_i \mathbf{w}_i^\top$ on mean-centered state $\mathbf{x}=\mathrm{vec}(\state-M)$ \citep{gao2024scaling}, minimizing
$\|\mathbf{x}-\sum_{i\in\mathrm{TopK}(\mathbf{a})}a_i\,\mathrm{vec}(\mathbf{v}_i\mathbf{w}_i^\top)\|^2+\lambda_{\mathrm{aux}}\mathcal{L}_{\mathrm{dead}}$,
where $\mathrm{TopK}(\mathbf{a})$ keeps the top-$k$ entries of $\mathbf{a}$ and $\mathcal{L}_{\mathrm{dead}}$ revives inactive atoms. The loss is reconstruction MSE plus the auxiliary dead-feature loss; no norm-matching term is used during training. The rank-1 decoder costs $d_k+d_v=256$ parameters per atom against $d_kd_v=16{,}384$ for a \flatsae dense atom (App.~\ref{app:encoderswap}). A flat atom can reconstruct the state but may mix several writes into one feature, making a single-write replacement test harder to interpret. We use \writesae for the rank-1 decoder matched to the model update; \bilinearsae denotes the matched-filter encoder variant used in the 4B generation probe. The training corpus is $5{,}000$ OpenWebText \citep{Gokaslan2019OpenWeb} sequences of length $1{,}024$ run through Qwen3.5-0.8B \citep{qwen2025qwen3} at layers $1$, $9$, and $17$, with an $80/20$ split. Atoms whose decoded direction matches a native rank-1 write are \emph{registers}; the rest are \emph{bundles}. \Sref{sec:mechanism} validates this observational partition by GMM, class-swap controls, and seed-stable counts.

\paragraph{Replacement protocol.}
At firing $(t,\ell,h)$, let $i^\star{=}\arg\max_i a_i(t)$ be the atom with the largest TopK coefficient. The replacement protocol swaps the native write $\Delta_{\mathrm{nat}}{=}\beta_t\mathbf{k}_t\mathbf{v}_t^\top$ for $\Delta_{\mathrm{atom}}{=}a_{i^\star}(t)\mathbf{v}_{i^\star}\mathbf{w}_{i^\star}^\top$, updates $S_t^{\ell,h} \!\mapsto\! S_t^{\ell,h} \!-\! \Delta_{\mathrm{nat}} \!+\! \Delta_{\mathrm{atom}}$, and continues the forward pass. The score is $\mathrm{KL}(p_{\mathrm{patched}}\,\|\,p_{\mathrm{baseline}})$ on the final output distribution, the metric \citet{zhang2023patching} recommend over logit-diff or accuracy for local activation patches.\label{alg:cachepatch}

\section{Experiments}
\label{sec:experiments}

We test three claims: the learned atoms split into two observable classes, individual atoms can replace model writes, and replacement works best when the atom rank matches the model's write rank. The causal metric is KL divergence between the final output distribution of the modified forward pass and the baseline.

\paragraph{Setup.}
We train \writesae on cached \gdn states from Qwen3.5-0.8B. The training set is $5{,}000$ OpenWebText \citep{Gokaslan2019OpenWeb} passages, sweeping layers $L \in \{1, 9, 17\}$ and heads $H \in \{0..15\}$. A cell is one $(\mathrm{layer}, \mathrm{head})$ pair. We use L9 H4 as the primary cell because a within-L9 sweep showed the largest separation between the two cosine components; the same ordering, atom better than deletion, then holds across the full layer (per-head distribution in \Sref{sec:mechvalid}). Cross-layer and cross-architecture extensions (DeltaNet, Mamba-2, GLA, Qwen3.5-4B/27B) follow in \Sref{sec:archscope}. The replacement test compares three conditions after one cache write: the SAE atom at its learned coefficient, deleting the write, and a random atom with the same coefficient. All KL values reported below are at the final output distribution. Partition statistics come from a two-component Gaussian mixture on median cosine with the native write. The full protocol is in App.~\ref{app:repro}.

\begin{figure*}[!t]
  \centering
  \includegraphics[width=0.85\textwidth]{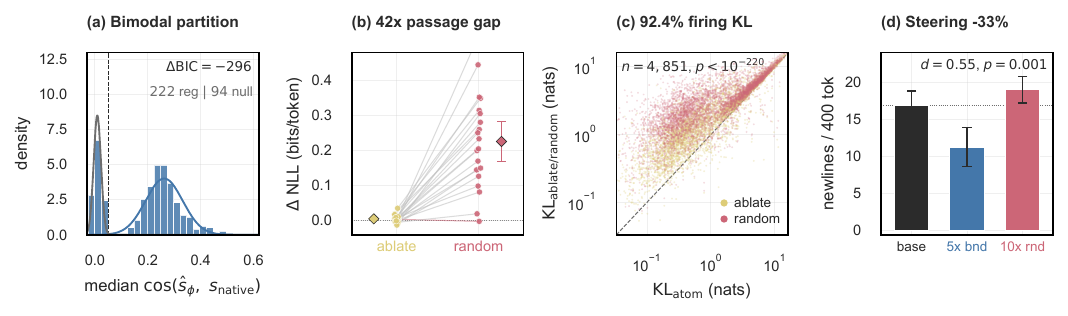}
  \caption{\textbf{Learned cache-write atoms produce lower KL at the final output distribution than deletion or random controls.} (a) Median cosine to the native write across the $316$ alive atoms; a two-component GMM separates them into $222$ registers and $94$ bundles. (b) On $20$ held-out OpenWebText passages, deleting every register firing costs $+0.005$ bits/token of passage NLL; a matched-norm random rank-1 write costs $+0.226$. (c) KL pooled across L1/L9/L17 evaluated positions, $n{=}4{,}851$, with the atom beating deletion on $92.4\%$. (d) Held-out Qwen3.5-4B generation probe: amplifying features more active at sentence boundaries reduces newlines from $16.8$ to $11.2$ ($p{=}0.001$, $d{=}0.55$).}
  \label{fig:causal_dashboard}
\end{figure*}

\subsection{Feature classes}
\label{sec:classes}
\label{sec:scale}

\writesae at Qwen3.5-0.8B L9 H4 trains $2{,}048$ atoms; $316$ survive on the validation split. A two-component Gaussian mixture on median cosine-to-native-write returns $222$ registers (mean cosine $0.26$) and $94$ bundles against $1{,}732$ null atoms, with Bayesian information criterion $\Delta\mathrm{BIC}=-296$ over the one-component null (Figure~\ref{fig:causal_dashboard}a). Cosine is the only feature used for this partition. Class membership is descriptive: bundles substitute on $89.0\%$ vs registers at $91.4\%$ at population scale, Mann-Whitney $p{=}0.24$ (App.~\ref{app:population_kl}). Here population scale means a per-atom substitution test over alive features, averaged across firing contexts. The causal probes in \Sref{sec:mechvalid} evaluate the alive population on axes excluded from the partition. The bundle mode is not the dense-SAE-latents phenomenon of \citet{sun2025densefeat}; App.~\ref{app:residual_sae_comparison} gives the comparison.

Three checks keep the observational partition separate from the causal tests. A random rank-1 control leaves top-$K$ overlap near $1$ across $47/48$ cells (Fig.~\ref{fig:selectivity}), the logit-change formula predicts effects at $R^2{=}0.98$, and learned atoms beat deleting the write across alive atoms at $89.8\%$ (App.~\ref{app:population_kl}). Seed runs reproduce the partition at coefficient of variation $4$--$12\%$ in counts and agree on $<1\%$ of specific atoms at cosine $>0.9$ \citep{paulo2025seedfeat}. Role counts are stable, but atom identities are seed-specific.

\begin{table}[t]
\centering
\scriptsize
\setlength{\tabcolsep}{3pt}
\caption{\textbf{Four seed-42 exemplar atoms used in the main text at Qwen3.5-0.8B L9 H4.} Activation rate orders the register rows; F1335 is the main register example because it fires on $5.24\%$ of validation tokens. Per-atom IDs are seed-specific; cross-seed identity rate is $<1\%$, while role classes reproduce. F87 is the bundle baseline in the native-norm control in \Sref{sec:mechvalid}.\protect\footnotemark}
\label{tab:exemplars}
\begin{tabular}{@{}llccccl@{}}
\toprule
Role exemplar (seed-42) & Role & Class & cos$_\mathbf{v}\uparrow$ & cos$_\mathbf{w}\uparrow$ & Fire rate & Top reader \\
\midrule
F1335 & delimiter gate       & register & \textbf{0.92}  & 0.42  & $5.24\%$ & L21 H10 at $7.5\times$  \\
F63   & factual-span register& register & \textbf{0.88}  & 0.63  & $2.69\%$ & L17 H4 at $5.2\times$   \\
F53   & proper-noun register & register & \textbf{0.99}  & 0.74  & $0.09\%$ & L5 H5 at $6.9\times$    \\
\midrule
F87   & bundle control       & bundle   & 0.02  & 0.01  & $15.9\%$ & substitution inverts    \\
\bottomrule
\end{tabular}
\end{table}
\footnotetext{F758 has the highest cosine alignment ($0.985$) but fires at activation rate $5\times 10^{-4}$, so its full context is in App.~\ref{app:cosine_free}.}

\paragraph{Exemplars and the register role.}
Table~\ref{tab:exemplars} uses F1335, F63, and F53 because they fire on natural text and read into different downstream cells. F1335 fires at delimiters next to list numerals, F53 on BPE sub-pieces of just-introduced proper nouns, F63 on factual-span continuations (Fig.~\ref{fig:feature_gallery} in App.~\ref{app:mech_figures} shows top-firing snippets, intervention KL, and reader enrichment for each). Pairwise Jaccard at exact tokens averages $0.001$ across the top ten registers, giving different surface triggers under similar write geometry. Independent seeds reproduce the partition at CV $4$--$12\%$ in atom counts, with $<1\%$ of specific atoms matching at cosine $>0.9$ across seeds.

\subsection{Mechanism validation}
\label{sec:mechvalid}
\label{sec:mechanism}

Substitution is a stronger criterion than reconstruction. The atom must replace the native write in the cache and preserve the downstream behavior. Across $20$ held-out OpenWebText passages at L9 H4, deleting every register firing raises NLL by $+0.005$ bits/token, while matched-norm random rank-1 writes raise it by $+0.226$, a $41.87\times$ gap that holds in $19/20$ passages (Figure~\ref{fig:causal_dashboard}b). The probes below ask whether the gap holds for single writes, across the two atom classes, and for logit changes predicted from the formula. Appendix~\ref{app:alternative-explanations} reports alternative-explanation controls.

\paragraph{Partition.}
The two-component split (Figure~\ref{fig:causal_dashboard}a) is observational at firing level: bundles substitute almost as well as registers at population scale (App.~\ref{app:population_kl}), and matched-norm random-rank-1 selectivity holds at $0.9953$ across $47/48$ cells.\footnote{Null-cosine median $0.00136$. BIC($k{=}2$) $= -679.18$ and BIC($k{=}3$) $= -683.33$; the marginal $\Delta\mathrm{BIC}{=}-4.15$ is too small to change the two-component operational separator we report.} Substitution performance therefore belongs to the alive dictionary population as a whole, beyond the cosine partition.

\begin{figure*}[t]
  \centering
  \includegraphics[width=0.85\textwidth]{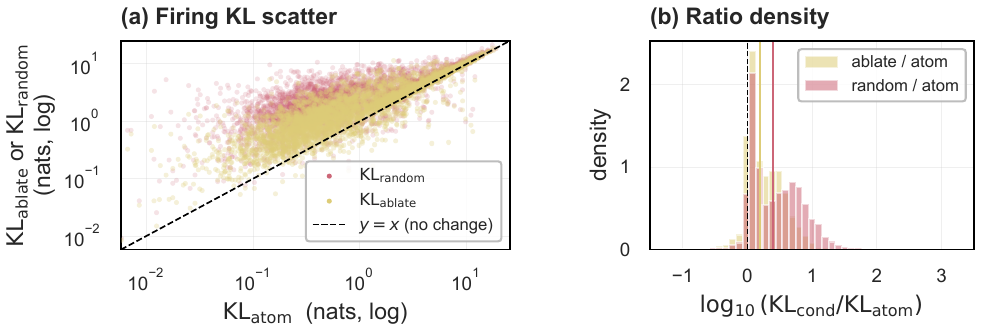}
  \caption{\textbf{Atom substitution beats both controls on $92.4\%$ of $n{=}4{,}851$ evaluated positions at L1/L9/L17 H4.} Left: log-log scatter of $\mathrm{KL}_{\text{delete}}$ (red) and $\mathrm{KL}_{\text{random}}$ (green) against $\mathrm{KL}_{\text{atom}}$, with $y{=}x$ for reference. Both distributions are above the identity line, and the strict chain $\mathrm{atom}<\mathrm{delete}<\mathrm{random}$ holds on $89.5\%$ of firings. Right: density of $\log_{10}(\mathrm{KL}_{\text{cond}}/\mathrm{KL}_{\text{atom}})$. The median ratio at each evaluated position is $1.55\times$ for deletion and $2.52\times$ for random, a $1.6\times$ separation of the controls; Table~\ref{tab:kl_triple} reports the ratio of the per-column medians ($1:2.70:4.90$).}
  \label{fig:kl_scatter}
\end{figure*}

\paragraph{Single-write replacement.}
At each firing we run three forward passes: the SAE atom $a_i(t) \cdot \mathbf{v}_i \mathbf{w}_i^\top$ replaces the native $\beta_t\,\mathbf{k}_t \mathbf{v}_t^\top$ write at position $t$, where $i$ has the largest TopK coefficient; the deletion pass removes the write; the random pass draws a fresh atom and uses the same coefficient $a_i(t)$. The learned atom beats deleting the write on $\mathbf{92.4\%}$ of $n{=}4{,}851$ firings, Wilson $95\%$ CI $[91.6, 93.1]$ (Fig.~\ref{fig:kl_scatter}). Cluster-bootstrap by feature widens that to $[90.91, 93.94]$.\footnote{$5{,}000$ resamples; the passage-clustered CI is $[90.90, 93.39]$ over $164$ clusters.} L1, L9, and L17 rates are $93.9\%$, $91.2\%$, and $92.3\%$, with Cliff's $\delta{=}+0.825$ at L9 (paired Wilcoxon $p < 10^{-200}$). The strict chain $\mathrm{KL}_\text{atom}<\mathrm{KL}_\text{delete}<\mathrm{KL}_\text{random}$ holds on $89.5\%$ of firings, so atom direction matters beyond write removal. In the L9 H4 population test over $87$ atoms, the atom beats deletion on $89.8\%$ of firings on average, $95\%$ CI $[88.1, 91.3]$. Bundle atoms ($n{=}57$) have mean $89.0\%$ and register atoms ($n{=}30$) have mean $91.4\%$, a $+2.4$pp gap that Mann-Whitney does not split at $p{=}0.24$. Replacement works for both cosine classes in the alive dictionary.

\paragraph{All L9 heads.}
The $92.4\%$ result pools evaluated positions across L1/L9/L17 H4. To check head selection, we repeat the replacement test on every L9 head with firings ($15/16$; H12 is dead), giving a mean atom-beats-deletion rate of $\mathbf{89.3\% \pm 2.6\%}$, range $82.6\%$--$93.2\%$. L9 H4 sits at $90.8\%$ on the L9-only pool, $+0.59\sigma$ above the L9 head mean. Per-head numbers and a strip plot are in App.~\ref{app:all16}, Fig.~\ref{fig:all16_strip}.

\begin{table}[t]
\centering
\small
\caption{\textbf{Atom substitution gives the lowest median KL at the final output distribution in every layer; the atom beats deleting the write on more than $91\%$ of positions per layer.} Each of the $n{=}4{,}851$ evaluated positions at Qwen3.5-0.8B head~4 contributes one triple of KL values: the atom at its firing coefficient, deleting the write, and a random atom scaled by the same coefficient. Cells report median $\pm$ MAD$/\sqrt{n}$; row-winners are bold. The ratio summary is the ratio of the per-column medians.}
\label{tab:kl_triple}
\begin{tabular}{lrccc c}
\toprule
Layer & $n$ & $\mathrm{KL}_{\text{atom}}\downarrow$ & $\mathrm{KL}_{\text{delete}}\downarrow$ & $\mathrm{KL}_{\text{random}}\downarrow$ & atom wins \\
\midrule
L1  & 1{,}500 & $\mathbf{2.07 \pm 0.06}$ & $2.82 \pm 0.07$ & $3.58 \pm 0.08$ & \textbf{93.9\%} \\
\rowcolor{black!5}
L9  & 1{,}851 & $\mathbf{0.40 \pm 0.02}$ & $1.09 \pm 0.04$ & $2.02 \pm 0.05$ & \textbf{91.2\%} \\
L17 & 1{,}500 & $\mathbf{1.82 \pm 0.06}$ & $2.58 \pm 0.07$ & $3.43 \pm 0.08$ & \textbf{92.3\%} \\
\midrule
\multicolumn{6}{l}{\emph{KL ratio (atom : deletion : random)} = $1 : 2.70 : 4.90$} \\
\bottomrule
\end{tabular}
\end{table}

\paragraph{Logit-change formula.}
Per-feature median $\mathrm{KL}$ across alive register atoms is $4\text{--}7\times 10^{-4}$, well below the random control.\footnote{Top-1 match $1.00$, per-feature KL $7\text{--}120\times$ tighter than random; CV $6.5\%/9.0\%$ across seven registers and bundle F87. We deep-copy the cache per condition because Qwen3.5's \gdn mutates state in place.} Pooled across firings the median is $0.40$ (Table~\ref{tab:kl_triple}). The triple $\mathrm{KL}_{\text{atom}}:\mathrm{KL}_{\text{delete}}:\mathrm{KL}_{\text{random}} = 1:2.70:4.90$ holds at L1 and L17. Eq.~\eqref{eq:closed_form} predicts the logit change at one token from gates the model already computes, with no fitted parameters, and obtains median per-feature $R^2 = 0.98$ across seven registers and bundle F87 (App.~\ref{app:closed_form}). The cosine factor accounts for the substitution gap; the output-embedding projection is not the limiting factor.

\paragraph{F87 fails when forced to native norm.}
F87 inverts when we amplify it to the native Frobenius norm. KL rises to $13\times$ the deletion condition, while register substitution at the same norm remains below the deletion floor.\footnote{F87 at cosine $0.01$: median $\mathrm{KL}\,7.0\times 10^{-3}$ vs deletion $5.0\times 10^{-4}$; top-1 swap on $7\%$ of firings.} The two atoms differ only in their cosine to the native write. F87's natural firing amplitude is small, so the population test cannot see the gap, and at natural amplitude F87 substitutes at $93.3\%$, indistinguishable from a register. The partition itself reappears at L1 ($\Delta\mathrm{BIC}{=}-6{,}774$) and L17 ($-390$); across SAE seeds, counts move at CV $4\text{--}12\%$ while $\approx 1\%$ of atoms reach cosine $>0.9$ across seeds (orthogonal-control check in App.~\ref{app:selectivity_sanity}).

\paragraph{Rank-2 decoder control.}
A rank-2 atom $A_i = \mathbf{v}_i^{(1)}\mathbf{w}_i^{(1)\top} + \mathbf{v}_i^{(2)}\mathbf{w}_i^{(2)\top}$ doubles parameters per entry. At all-$16$-head L9 substitution, rank-2 changes perplexity by $+0.82\%$ against $+0.76\%$ for rank-1, a $\Delta{=}+0.06$pp parity result (App.~\ref{app:rank2_perhead}). Because \gdn writes one rank-1 outer per step, rank-2 atoms do not improve the single-write replacement metric.

\subsection{Architectural scope}
\label{sec:archscope}

\begin{figure*}[t]
  \centering
  \includegraphics[width=0.85\textwidth]{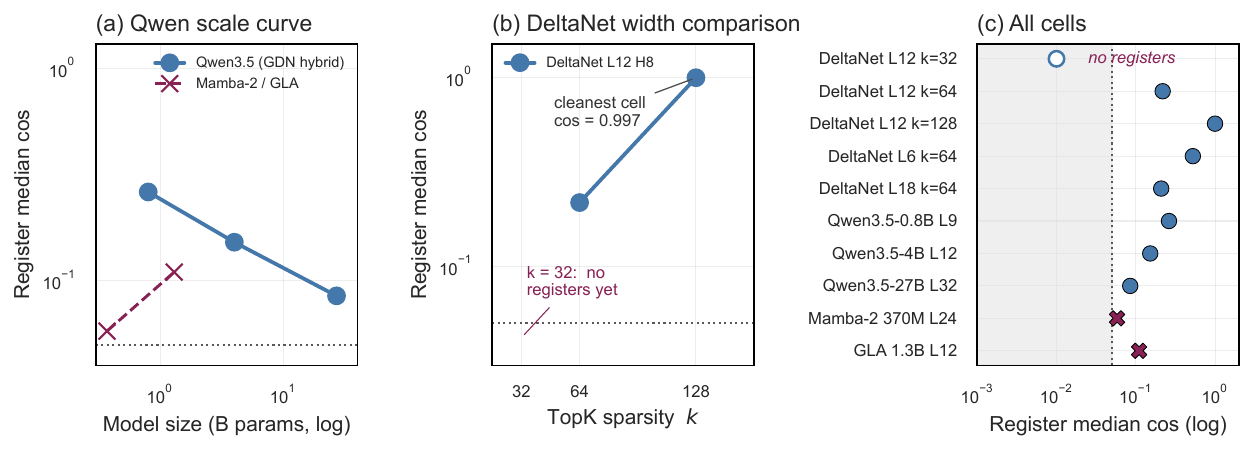}
  \caption{\textbf{Write rank separates the tested cells by register-cosine separation (KS $p{=}1.2\times 10^{-10}$).} (a) Register median cosine down the Qwen3.5 ladder runs $0.262$ (0.8B), $0.152$ (4B), $0.085$ (27B); Mamba-2 and GLA at matched scale stay below the $0.05$ threshold. (b) DeltaNet L12 H8 over TopK sparsity: no register-class atoms at $k{=}32$, peak $0.997$ at $k{=}128$. (c) All ten cells on a single log axis. Blue points are outer-product writes; red points are diagonal or scalar-gated states.}
  \label{fig:arch_boundary}
\end{figure*}

Eq.~\eqref{eq:closed_form} predicts that write rank, not parameter count, governs how closely atoms align with native writes. Five host architectures test the prediction. GDN and DeltaNet \citep{yang2024deltanet} write rank-1 outers, RWKV-7 \citep{peng2025rwkv7} writes rank-2, and Mamba-2 \citep{dao2024mamba2} and GLA \citep{yang2023gla} update a diagonal state. Softmax attention is outside the scope; the variable here is the recurrent write rule. The partition appears across the $34\times$ Qwen3.5 scale range and a five-cell DeltaNet sparsity sweep (Figure~\ref{fig:arch_boundary}).

\paragraph{DeltaNet and scale.}
DeltaNet L12 H8 at $k{=}128$ has the largest register/null separation we measured: register median cosine $0.997$ and register/null ratio $383\times$. That cell runs with \texttt{use\_gate=false}, so the update is purely bilinear in $\mathbf{k}_t \mathbf{v}_t^\top$; Qwen3.5 hybrids use the convex gate that DeltaNet drops. The Qwen3.5 cosine ladder reads $0.262$ at 0.8B, $0.152$ at 4B, and $0.085$ at 27B (App.~\ref{app:scaling}, Fig.~\ref{fig:scale_saturation}), with register counts of $220$ and $147$ at 4B and 27B. Qwen3.5-27B is $11.7\times$ below the DeltaNet cell even though both write rank-1 outers, consistent with the gate difference between them. Causal substitution at Qwen3.5-4B L12 H8 came out at chance under the same SAE recipe, a known training-objective gap (\Sref{sec:falsification}) rather than an architecture failure.

\paragraph{Host-matched WriteSAEs on Mamba-2 and RWKV-7.}
Each host architecture uses a WriteSAE decoder matching its native write rule.\footnote{$n_\mathrm{feat}{=}2{,}048$. RWKV-7 register max cosine $0.671$. GLA scalar-gated bilinear gives register median $0.110$ \citep{yang2023gla, hu2025comba}. Residual atoms cannot occupy the cache slot (App.~\ref{app:residual_sae_comparison}).} The observed register-cosine ordering is GDN ($0.262$) $>$ RWKV-7 ($0.180$) $>$ Mamba-2 ($0.0575$). Mamba-2-370M L24 H0 has $217$ register atoms against $1{,}831$ null; RWKV-7-1.5B L12 H0 has $200$ register atoms against $541$ null. The firing-level KS test uses cluster-bootstrap by feature with Holm correction over the four pairwise contrasts. GDN-Mamba-2 and DeltaNet-Mamba-2 clear $p_{\text{Holm}}<10^{-6}$; the within-rank-1 DeltaNet-GDN comparison does not separate at $\alpha{=}0.05$, as expected when the only difference is gate strength. Cross-architecture crosscoders \citep{jiralerspong2026crossarch} and feature universality \citep{lan2024univsae} extend the write-rule question beyond residual-stream features.

\paragraph{Mamba-2 replacement test.}
At Mamba-2-370M L24 H0, we replace the native diagonal update $\mathrm{d}t \cdot \mathrm{diag}(\mathbf{B}_t) \cdot \mathbf{x}_t$ with a WriteSAE atom $a_i \cdot \mathbf{v}_i$, where $\mathbf{v}_i$ is the SAE's diagonal decoder atom and $a_i$ is its activation. Atom beats deleting the write on $\mathbf{88.08\%}$ of $n{=}2{,}500$ firings drawn from $100$ atoms (60 register, 40 bundle by cosine partition), Wilson $95\%$ CI $[86.8, 89.3]$. Median KL is $0.97$ for the atom, $1.62$ for deletion, and $2.32$ for a random atom scaled by the same coefficient. The random control has $2.4\times$ higher KL than the atom; register and bundle are indistinguishable at Mann-Whitney $p{=}0.76$. Per-atom win rate is uncorrelated with cosine to the native write (Pearson $r{=}0.008$, $p{=}0.93$), matching the 0.8B GDN pattern. The averaged replacement test now succeeds in both GDN ($89.8\%$ at $87$ atoms) and Mamba-2 ($88.1\%$ at $100$ atoms). The $92.4\%$ result is the single-write replacement rate at one GDN head; the cross-host ordering GDN $>$ RWKV-7 $>$ Mamba-2 supports the claim that the native write rule matters.

\subsection{Ablations}
\label{sec:ablations}

\paragraph{Encoder and remaining ablations.}
At matched $\nf{=}2{,}048$, sparsity $k{=}32$, and training budget, \writesae's bilinear encoder yields $32\%$ dead features against \flatsae's $80\%$ across a $720$-run sweep (App.~\ref{app:encoderswap}), while BatchTopK and JumpReLU both recover the same register/bundle partition under the bilinear encoder (App.~\ref{app:sae_variants}). The encoder controls alive-feature count; the sparsity mechanism does not.

\paragraph{Probes, SVD, and SAE alternatives.}
Linear probes detect class membership but cannot substitute into the cache because the replacement must have the same rank-1 shape as the model write. PCA top-1 of writes is anti-correlated or near zero on every register exemplar (cosine $-0.216$, $-0.045$, $-0.075$ at F53, F63, F1335) while the SAE atom recovers the native write direction. The best-performing non-bilinear baseline in this sweep trains a flat TopK SAE on $\mathrm{vec}(S_t)$ and substitutes its top-1 SVD outer product. On Mamba-2-370M L24 H0 the architecture-matched decoder improves over flat-SAE-SVD by $+6.55$pp ($82.85\%$ vs $76.30\%$); on RWKV-7-1.5B L12 H0 both methods are near chance ($45.3\%$ vs $47.8\%$). On Qwen3.5-0.8B \gdn L9 H4 the two finish within $0.11$pp ($91.25\%$ vs $91.36\%$, $n{=}1{,}851$): gate decay already rank-1 dominates the state, so SVD top-1 of a flat-SAE atom recovers the direction the trained dictionary picks. The prior matters where the state is not rank-1 dominated by gating decay. Matching-pursuit SAE evaluation \citep{bussmann2025matchpursuit} reports similar host-agnostic ranking on transformer residuals; the substitution test here is architecture-aware.

\section{Cache Intervention Probes}
\label{sec:steering}

The closed-form expression of \Sref{sec:falsification} explains a firing; it also chooses one. We use it as the engine for three interventions on a single GDN layer-head: erasing an atom on its own firings, writing the formula's preferred direction at a single cache position, and writing that direction at three consecutive positions during decoding. A held-out 4B-scale probe then checks whether the same effects survive at scale. \emph{Target inclusion} is our metric for the generation probes: the predicted token appears at least once in the 20-token greedy continuation.

\paragraph{Cache-slot erasure.}
Erasing one atom's contribution affects only the token that atom promotes. Take F412 on Qwen3.5-0.8B L9 H4. At its $n{=}150$ natural firing positions, removing the rank-1 write from the cache drops the most-affected token's log-probability by a median $0.116$ nats (paired Wilcoxon $p{=}1.07\times 10^{-6}$, $95\%$ CI $[-0.265, -0.042]$). The same write at $n{=}150$ non-firing positions does not significantly change the logit (median $\Delta\log p{=}{+}0.016$ at $4\times a^\ast$, $p{=}0.15$): context masks writes at positions where the atom would not have chosen to fire. Dose-response, target, and rank tables are in App.~\ref{app:memory_edit}.

\paragraph{Single-position sign prediction.}
For each target token $T$, the closed form picks a unit-norm direction $v_T^\ast = W_O[\mathrm{head}]^\top W_U[T]/\|\cdot\|$ that should raise $T$'s logit when written into the cache. Sign holds on $84.6\%$ of $n{=}2{,}000$ trials ($95\%$ CI $[83.0, 86.2]$); magnitude is harder. Pearson correlation between predicted and measured shift comes out at $r{=}0.162$, with median measured-to-predicted ratio $1.08$ and pooled $R^2{=}{-}0.06$. Small noisy effects swamp the pool. None of that matters for greedy decoding, because the target only has to overtake the native top-$1$ at one step. Per-feature breakdowns are in App.~\ref{app:predictive_steering}.

\paragraph{Generation edit.}
One cache write often does not cross the bar. The target has to overtake the native top-$1$, and a single push can lose ground later. Three writes work better. Take tokens the unmodified model initially ranks between $100$ and $1000$ ($n{=}300$ contexts). Writing $v_T^\ast$ at three consecutive cache positions with magnitude $m{=}3.0\,\|\mathbf{k}_t\mathbf{v}_t^\top\|$ raises target inclusion from $33.3\%$ to $\mathbf{100\%}$ under greedy decoding; the median rank shift is $+517$.

Pooled across the four target strata, the rate moves from $8.3\%$ to $25.0\%$ over $1{,}200$ trials. Rank improves on $77.4\%$ of trials. First-step logp lift is $+1.27$ nats (Table~\ref{tab:behavioral_body}). Out-of-context targets shift rank but never reach top-$1$ inside $20$ tokens.

The dose response is non-monotone. $m{=}1.5$ reaches $66.7\%$ on mid-rank targets; $m{=}3.0$ reaches $100\%$; $m{=}6.0$ oversaturates back to $16.7\%$ pooled. Full breakdown in App.~\ref{app:behavioral_steering}.

\begin{figure}[!htbp]
  \centering
  \includegraphics[width=0.36\linewidth]{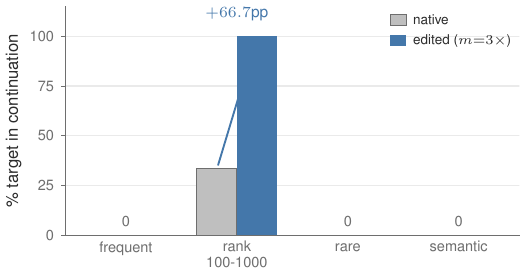}
  \caption{\textbf{Writing the direction chosen by the formula at three cache positions raises target inclusion from $33.3\%$ to $100\%$ for targets initially ranked $100$--$1000$ by the unmodified model ($n{=}300$).} Target inclusion by class at $m{=}3\times$ on Qwen3.5-0.8B L9 H4; native (gray) vs edited cache (atom-blue). Out-of-context targets shift rank but remain at $0\%$.}
  \label{fig:behavioral_steering_body}
\end{figure}

\begin{table}[!htbp]
\centering
\scriptsize
\setlength{\tabcolsep}{5pt}
\caption{\textbf{Directions chosen by the formula increase target inclusion in a constrained generation probe.} Pooled rows include frequent targets, targets initially ranked $100$--$1000$ by the unmodified model, rare targets, and semantic targets at Qwen3.5-0.8B L9 H4 (Fig.~\ref{fig:behavioral_steering_body}; App.~\ref{app:behavioral_steering}).}
\label{tab:behavioral_body}
\begin{tabular}{lcccc}
\toprule
Target set & $n$ & target in continuation & native & first-step lift \\
\midrule
Pooled & 1{,}200 & $25.0\%$ & $8.3\%$ & $+1.27$ nats \\
\textbf{Ranks $100$--$1000$} & 300 & $\mathbf{100\%}$ & $33.3\%$ & $+1.27$ nats \\
\bottomrule
\end{tabular}
\end{table}

\paragraph{Newline-rate edit on a held-out 4B model.}
The same dictionary transfers to a larger host: Qwen3.5-4B-Base at layer~9, with no retraining. The boundary signal is concentrated. Score each feature by mean activation on sentence-boundary tokens minus mean activation on other tokens. The top-10 features in each of the 32 heads sit well above the rest of the pool.

Amplification adds a positive offset to those SAE coefficients during decoding and leaves the residual stream alone. The dose-matched control draws a random feature from the same pool. Newlines per generation is the primary readout; paragraph count and mean word length are surface-quality checks. Doses run $2\times$, $5\times$, and $10\times$ the mean boundary activation, with $400$ tokens at temperature~$0.7$ over $40$ prompts.

\begin{figure}[t]
  \centering
  \includegraphics[width=0.36\linewidth]{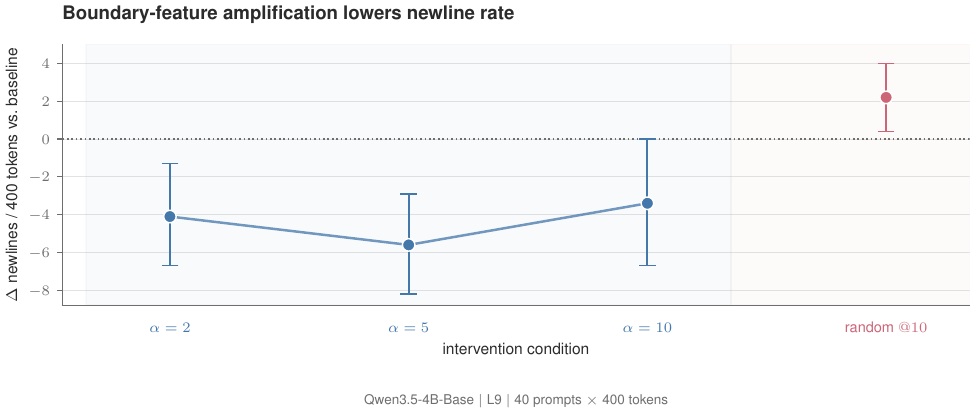}
  \caption{\textbf{Boundary-feature amplification changes newline rate in a held-out 4B probe.} Mean newlines per 400 generated tokens on Qwen3.5-4B-Base L9, $n{=}40$ prompts. Amplifying \bilinearsae features selected by sentence-boundary activation at $5\times$ changes the count from $16.8$ to $11.2$ ($-33\%$, $p{=}0.001$); the response saturates and rebounds toward baseline at $10\times$. The dose-matched random-feature control at $10\times$ changes the count in the opposite direction (above baseline). Word-length stays within $\pm 0.3$ characters across conditions (Table~\ref{tab:steering}).}
  \label{fig:steering}
\end{figure}

\paragraph{Results.}
Amplifying those boundary features reduces line breaks. At the $5\times$ dose, mean newlines per $400$ tokens fall from $16.8$ to $11.2$. That is a $33\%$ reduction across $n{=}40$ prompts. Paired $t$-test $p{=}0.001$, Cohen's $d{=}0.55$ (Fig.~\ref{fig:steering}). The drop is direction-specific.

We selected the $5\times$ dose post hoc from the full $\{1\times, 2\times, 5\times, 10\times\}$ sweep, so the effect has to survive multiple-testing correction. Bonferroni across the four doses leaves the $5\times$ result at $p_\text{adj}{=}0.004$; the $2\times$ effect ($p_\text{raw}{=}0.015$) does not survive. The dose curve also saturates, so at $10\times$ the newline count climbs back to $13.4$.

Surface metrics move in the predicted direction at smaller amplitude. Paragraph count falls from $7.5$ to $6.2$. Mean word length shifts from $5.54$ to $5.24$ characters. The dose-matched random-feature control at $10\times$ takes newlines the other way, up to $19.0$, above the $16.8$ baseline. The boundary-feature direction is selective, not dose-driven.

\begin{table}[t]
\centering
\small
\caption{\textbf{Boundary-feature amplification reduces newline rate by $33\%$ on a held-out 4B model.} Generation metrics on Qwen3.5-4B-Base ($95\%$ bootstrap CIs, $n{=}40$ prompts, $400$ tokens each). Word length is reported in characters. The primary comparison is $5\times$ amplification vs.\ baseline.}
\label{tab:steering}
\begin{tabular}{llccc}
\toprule
Condition & Dose & Newlines & Paragraphs & Word length \\
\midrule
Baseline & 0 & 16.8\,[14.8, 18.9] & 7.5\,[6.6, 8.5] & 5.54 \\
Amplify & $2{\times}$ & 12.7\,[10.1, 15.5] & 6.3\,[5.2, 7.4] & 5.34 \\
Amplify & $5{\times}$ & 11.2\,[8.6, 13.9] & 6.2\,[4.9, 7.4] & 5.24 \\
Amplify & $10{\times}$ & 13.4\,[10.1, 16.8] & 7.4\,[5.9, 8.9] & 5.29 \\
Random & $10{\times}$ & 19.0\,[17.2, 20.8] & 7.8\,[7.0, 8.7] & 5.50 \\
\bottomrule
\end{tabular}
\end{table}

\paragraph{Controls.}
In a separate 0.8B experiment, \flatsae amplification reduces word length (4.86$\to$3.53) and leaves paragraph count unchanged: the intervention degrades surface quality without shifting document structure. Only \bilinearsae features produced the target newline reduction. \flatsae and \ranksae (a dense-encoder rank-1 SAE on $\state$) results come from separate 0.8B pilots with different feature pools and higher dead-feature rates ($80\%$ vs.\ $32\%$), so they serve as negative controls rather than matched comparisons. The amplification/suppression asymmetry has a structural explanation: TopK activations are nonnegative, so the sparse code has no negative loadings, and suppression can at most clamp a coefficient to zero while amplification can increase it.

A second cell on the same 4B model shows the constraint behind the layer-9 newline result. We repeat the protocol at L12 H8 with a different target behavior: amplifying the top-$10$ features more active on proper nouns than on other tokens at $5\times$ on $40$ prompts of $150$ tokens. The capitalized-word rate moves from $0.0862$ at baseline to $0.0852$, a $\Delta{=}-0.001$ shift at $d{=}-0.03$, $p{=}0.86$ against baseline (reported failure) and $p{=}0.18$ against a dose-matched random-feature control. The top-10 proper-noun features at L12 H8 have a maximum $|\mathrm{mean\_diff}|{=}0.0047$ between proper-noun and other tokens, an order of magnitude below the layer-9 boundary signal. The L12 H8 dictionary does not contain an atom that separates proper nouns from other tokens, and the signal did not transfer in this cell. Cache generation interventions require an atom whose activation separates the target behavior; this requirement is not automatic across cells.

\section{Related Work}
\label{sec:related}

\paragraph{Sparse autoencoders, circuits, and causal edits.}
Transformer SAEs decode the residual stream \citep{elhage2022toymodels, bricken2023monosemanticity, gao2024scaling}, so their atoms are vectors. The closest non-residual line is transcoders \citep{dunefsky2024transcoders, paulo2025skiptranscoder, marks2024feature}: a transcoder substitutes an MLP's contribution, but that contribution is still a vector. \writesae sits at a different intervention site, where the layer writes a matrix; this forces atoms with matching matrix shape. Dictionary learning outside ML has had matrix atoms for decades \citep{ravishankar2015soupdil, olshausen1996sparse}. A long line of SAE-side variants has changed objectives and decoder families without leaving the residual-stream target \citep{makhzani2013ksparse, matryoshkasae, dooms2025bilinear, koromilas2026polysae, engels2024notlinear}. Causal evaluations of these dictionaries \citep{karvonen2025saebench, gurnee2023sparseprobing, mueller2025mib, lieberum2024gemmascope} and the activation-editing line \citep{conmy2023acdc, geiger2024das, meng2023memit, wu2024reft, axbench, ameisen2025circuittracing, lindsey2025biology} give us our protocol; the difference is that our test perturbs one rank-1 cache write rather than a residual vector. Concurrent rank-sparse attention work \citep{he2025lorsa} asks how rank shapes attention's own decomposition; for us, rank is set by the host write rule.

\paragraph{Matrix-recurrent states and low-rank structure.}
Each state-space and hybrid recurrent layer adds something low-rank to its cache every token. Fast-weight programmers \citep{schmidhuber1992fwp, schlag2021fwp} introduced the rank-1 outer-product update; linear attention \citep{katharopoulos2020linear} and test-time-training layers \citep{sun2024ttt} reuse the same arithmetic; modern hybrids scale it across RetNet, GLA, \gdn, DeltaNet, RWKV-7, Mamba-2, Hedgehog, and mixture-of-memory variants \citep{sun2023retnet, yang2023gla, yang2024gdn, yang2024deltanet, peng2025rwkv7, dao2024mamba2, zhang2024hedgehog, du2025mom}. Prior probes of the recurrent state describe its content \citep{pitorro2025latim, airlangga2025mamba, arora2025ssmevals, okpekpe2025recall, sharma2024mamba} or compress it after the fact \citep{chang2025xkv, nazari2026rank}. We target the per-write structure that produced the state instead, by substituting a learned rank-1 atom for the layer's own write.

The shape of an SAE atom has to match the shape of what the host layer writes. SAE-side and adaptation-side work usually probes residual streams or hidden-state offsets \citep{paulo2024rnntransfer, wang2024universality, hossain2025mambamemory, sunkumohan2026asb, yap2026steeringmoe, galim2024peftssm}, while parameter-side decompositions sparsify weights or Jacobians upstream of the cache \citep{farnik2025jacobian, braun2025apd, bushnaq2025spd}. We match the atom rank to the host write rule: rank-1 for outer-product writes, diagonal for Mamba-2, and rank-2 for RWKV-7. Mamba-3 \citep{lahoti2026mamba3} adds new state-input geometries (exponential-trapezoidal, complex, MIMO) that would test the same rule. Our observed GDN $>$ RWKV-7 $>$ Mamba-2 ordering aligns with the expressivity gap that outer-product corrections open over diagonal-only updates \citep{siems2025deltaproduct}.

\section{Discussion and Conclusion}
\label{sec:discussion}

The main claim is limited but concrete: when the model writes a rank-1 matrix update, a rank-1 SAE atom can often replace that single update. GDN shows the clearest register class and the best fit to the closed-form logit shift ($R^2{=}0.98$). Mamba-2 and RWKV-7 show weaker alignment, but learned atoms still beat deleting the write at population scale. The replacement test transfers across recurrent families; the GDN gate coefficient does not.

\subsection{Limitations and future work}
\label{sec:scope}
We focus on Qwen3.5-0.8B GDN L9 H4 because its native write is a single rank-1 outer product, the same shape used by \writesae atoms. Per-atom identity varies across SAE seeds (less than $1\%$ of atoms match at cosine $>0.9$); the register/bundle class is the unit that transfers across seeds. The reported failures below point to a write-aligned loss for 4B, a Mamba-2-specific logit formula, substitutions that handle several active atoms at once, and Mamba-3 atom shapes \citep{lahoti2026mamba3}. Scaling F412 erasure to multi-feature edits remains open.

\subsection{Falsification and observed failures}
\label{sec:falsification}

Four reported failures bound the result: the GDN logit formula does not transfer directly to Mamba-2, the same SAE recipe does not give a 4B firing-level effect, rank-2 atoms give only a small top-1 gain, and the Mamba-2 generation edit fails.

\paragraph{Logit formula outside GDN.}
The formula $\Delta\ell \approx G \cdot \langle \mathbf{w}_i, \mathbf{q}_t \rangle \cdot \langle \mathbf{v}_i, W_U[\mathrm{tok}] \rangle$ predicts measured effects at $R^2{=}0.98$ at GDN L9 H4. On Mamba-2 L24 H0 and Qwen3.5-4B L12 H8, the same formula yields negative $R^2$ ($-0.07$ and $-0.05$). Architectures without the same multiplicative gate need their own coefficient form. The partition test and cache replacement transfer across host architectures; the gate coefficient does not.

\paragraph{Larger Qwen model.}
At Qwen3.5-4B L12 H8 with the same SAE recipe used at 0.8B, atoms beat deleting the write only at chance ($48\%$ pooled, $n{=}600$). The 4B SAE reaches better validation MSE than the 0.8B SAE ($5.6\times 10^{-6}$ vs $2.2\times 10^{-5}$), so the failure is not under-training. Reconstruction optimizes state recovery; substitution requires write-direction alignment. At scale these objectives decouple, motivating a write-aligned training objective.

\paragraph{Rank-2 top-1 comparison.}
Independent rank-1 and rank-2 SAEs use different feature-index conventions. A head-to-head test using each SAE's top-1 atom per firing gives rank-2 a small edge (Cliff's $\delta{=}-0.028$, $p<10^{-71}$, $n{=}24{,}600$), with median log-ratio of KL $-0.035$. Because GDN writes one rank-1 outer per step, the practical gain is small when replacing one cache write.

\paragraph{Mamba-2 generation edit.}
Applying the GDN-derived generation edit to Mamba-2 L24 H0 produces no lift in target inclusion across $n{=}3{,}600$ trials, with median first-step logp lift ${\approx}\,0$. This failure matches the logit-formula failure outside GDN: the direction is approximate at Mamba-2 and too small to alter greedy generation. Together with the GDN success, it scopes this cache edit to architectures with the same gate and readout structure.

\newpage
\bibliographystyle{plainnat}
\bibliography{references}

@article{bricken2023monosemanticity,
  title={Towards Monosemanticity: Decomposing Language Models With Dictionary Learning},
  author={Bricken, Trenton and Templeton, Adly and Batson, Joshua and Chen, Brian and Jermyn, Adam and Conerly, Tom and Turner, Nick and Anil, Cem and Denison, Carson and Askell, Amanda and Lasenby, Robert and Wu, Yifan and Kravec, Shauna and Schiefer, Nicholas and Maxwell, Tim and Joseph, Nicholas and Hatfield-Dodds, Zac and Tamkin, Alex and Nguyen, Karina and McLean, Brayden and Burke, Josiah E. and Hume, Tristan and Carter, Shan and Henighan, Tom and Olah, Christopher},
  journal={Transformer Circuits Thread},
  year={2023},
  url={https://transformer-circuits.pub/2023/monosemantic-features}
}

@inproceedings{cunningham2024saes,
  title={Sparse Autoencoders Find Highly Interpretable Features in Language Models},
  author={Cunningham, Hoagy and Ewart, Aidan and Riggs, Logan and Huben, Robert and Sharkey, Lee},
  booktitle={International Conference on Learning Representations},
  year={2024},
  eprint        = {2309.08600},
  archivePrefix = {arXiv},
  primaryClass  = {cs.LG}
}

@article{gao2024scaling,
  title={Scaling and Evaluating Sparse Autoencoders},
  author={Gao, Leo and la Tour, Tom Dupr{\'e} and Tillman, Henk and Goh, Gabriel and Troll, Rajan and Radford, Alec and Sutskever, Ilya and Leike, Jan and Wu, Jeffrey},
  year={2025},
  eprint={2406.04093},
  archivePrefix={arXiv},
  primaryClass={cs.LG},
  url={https://arxiv.org/abs/2406.04093},
  journal = {arXiv preprint arXiv:2406.04093}
}

@article{templeton2024scaling,
  title={Scaling Monosemanticity: Extracting Interpretable Features from {Claude} 3 {Sonnet}},
  author={Templeton, Adly and Conerly, Tom and Marcus, Jonathan and Lindsey, Jack and Bricken, Trenton and Chen, Brian and Pearce, Adam and Citro, Craig and Ameisen, Emmanuel and Jones, Andy and Cunningham, Hoagy and Turner, Nicholas L. and McDougall, Callum and MacDiarmid, Monte and Freeman, C. Daniel and Sumers, Theodore R. and Rees, Edward and Batson, Joshua and Jermyn, Adam and Carter, Shan and Olah, Chris and Henighan, Tom},
  journal={Transformer Circuits Thread},
  year={2024},
  url={https://transformer-circuits.pub/2024/scaling-monosemanticity}
}

@article{rajamanoharan2024gated,
  title={Improving Dictionary Learning with Gated Sparse Autoencoders},
  author={Rajamanoharan, Senthooran and Conmy, Arthur and Smith, Lewis and Lieberum, Tom and Varma, Vikrant and Kram{\'a}r, J{\'a}nos and Shah, Rohin and Nanda, Neel},
  year={2024},
  eprint={2404.16014},
  archivePrefix={arXiv},
  primaryClass={cs.LG},
  url={https://arxiv.org/abs/2404.16014},
  journal = {arXiv preprint arXiv:2404.16014}
}

@article{rajamanoharan2024jumprelu,
  title={Jumping Ahead: Improving Reconstruction Fidelity with {JumpReLU} Sparse Autoencoders},
  author={Rajamanoharan, Senthooran and Lieberum, Tom and Sonnerat, Nicolas and Conmy, Arthur and Varma, Vikrant and Kram{\'a}r, J{\'a}nos and Nanda, Neel},
  year={2024},
  eprint={2407.14435},
  archivePrefix={arXiv},
  primaryClass={cs.LG},
  url={https://arxiv.org/abs/2407.14435},
  journal = {arXiv preprint arXiv:2407.14435}
}

@misc{lindsey2025biology,
  title={On the Biology of a Large Language Model},
  author={Lindsey, Jack and Gurnee, Wes and Ameisen, Emmanuel and Chen, Brian and Pearce, Adam and Turner, Nicholas L. and Olah, Chris and Batson, Joshua},
  year={2025},
  month={mar},
  howpublished={Transformer Circuits Thread},
  url={https://transformer-circuits.pub/2025/attribution-graphs/biology.html}
}

@misc{ameisen2025circuittracing,
  title={Circuit Tracing: Revealing Computational Graphs in Language Models},
  author={Ameisen, Emmanuel and Lindsey, Jack and Pearce, Adam and Gurnee, Wes and Turner, Nicholas L. and Chen, Brian and Citro, Craig and Abrahams, David and Carter, Shan and Hosmer, Basil and Marcus, Jonathan and Sklar, Michael and Templeton, Adly and Bricken, Trenton and McDougall, Callum and Cunningham, Hoagy and Henighan, Thomas and Jermyn, Adam and Jones, Andy and Persic, Andrew and Qi, Zhenyi and Thompson, T. Ben and Zimmerman, Sam and Rivoire, Kelley and Conerly, Thomas and Olah, Chris and Batson, Joshua},
  year={2025},
  month={mar},
  howpublished={Transformer Circuits Thread},
  url={https://transformer-circuits.pub/2025/attribution-graphs/methods.html}
}

@inproceedings{conmy2023acdc,
  title={Towards Automated Circuit Discovery for Mechanistic Interpretability},
  author={Conmy, Arthur and Mavor-Parker, Augustine and Lynch, Aengus and Heimersheim, Stefan and Garriga-Alonso, Adri{\`a}},
  booktitle={Advances in Neural Information Processing Systems (NeurIPS)},
  year={2023},
  eprint        = {2304.14997},
  archivePrefix = {arXiv},
  primaryClass  = {cs.LG}
}

@inproceedings{marks2024feature,
  title={Sparse Feature Circuits: Discovering and Editing Interpretable Causal Graphs in Language Models},
  author={Marks, Samuel and Rager, Can and Michaud, Eric J. and Belinkov, Yonatan and Bau, David and Mueller, Aaron},
  booktitle={International Conference on Learning Representations},
  year={2025},
  eprint        = {2403.19647},
  archivePrefix = {arXiv},
  primaryClass  = {cs.LG}
}

@inproceedings{geiger2024das,
  title={Finding Alignments Between Interpretable Causal Variables and Distributed Neural Representations},
  author={Geiger, Atticus and Wu, Zhengxuan and Potts, Christopher and Icard, Thomas and Goodman, Noah D.},
  booktitle={Conference on Causal Learning and Reasoning ({CLeaR})},
  year={2024},
  eprint        = {2303.02536},
  archivePrefix = {arXiv},
  primaryClass  = {cs.LG}
}

@article{paulo2024rnntransfer,
  title={Does Transformer Interpretability Transfer to RNNs?},
  author={Paulo, Gon{\c{c}}alo and Marshall, Thomas and Belrose, Nora},
  year={2024},
  eprint={2404.05971},
  archivePrefix={arXiv},
  primaryClass={cs.LG},
  url={https://arxiv.org/abs/2404.05971},
  journal = {arXiv preprint arXiv:2404.05971}
}

@inproceedings{hossain2025mambamemory,
  title={Characterizing {M}amba's Selective Memory using Auto-Encoders},
  author={Hossain, Tamanna and Logan IV, Robert L. and Jagadeesan, Ganesh and Singh, Sameer and Tetreault, Joel and Jaimes, Alejandro},
  booktitle={Findings of IJCNLP-AACL},
  year={2025},
  eprint        = {2512.15653},
  archivePrefix = {arXiv},
  primaryClass  = {cs.LG}
}

@article{wang2024universality,
  title={Towards Universality: Studying Mechanistic Similarity Across Language Model Architectures},
  author={Wang, Junxuan and Ge, Xuyang and Shu, Wentao and Tang, Qiong and Zhou, Yunhua and He, Zhengfu and Qiu, Xipeng},
  year={2025},
  eprint={2410.06672},
  archivePrefix={arXiv},
  primaryClass={cs.LG},
  url={https://arxiv.org/abs/2410.06672},
  booktitle={International Conference on Learning Representations (ICLR)},
  journal = {arXiv preprint arXiv:2410.06672}
}

@inproceedings{karvonen2025saebench,
  title={{SAEBench}: A Comprehensive Benchmark for Sparse Autoencoders in Language Model Interpretability},
  author={Karvonen, Adam and Rager, Can and Lin, Johnny and Tigges, Curt and Bloom, Joseph and Chanin, David and Lau, Yeu-Tong and Farrell, Eoin and McDougall, Callum and Ayonrinde, Kola and Till, Demian and Wearden, Matthew and Conmy, Arthur and Marks, Samuel and Nanda, Neel},
  booktitle={Proceedings of the 42nd International Conference on Machine Learning (ICML)},
  series={PMLR},
  volume={267},
  pages={29223--29264},
  year={2025},
  eprint        = {2503.09532},
  archivePrefix = {arXiv},
  primaryClass  = {cs.LG}
}

@article{dooms2025bilinear,
  title={Finding Manifolds With Bilinear Autoencoders},
  author={Dooms, Thomas and Gauderis, Ward},
  year={2025},
  eprint={2510.16820},
  archivePrefix={arXiv},
  primaryClass={cs.LG},
  url={https://arxiv.org/abs/2510.16820},
  journal = {arXiv preprint arXiv:2510.16820}
}

@article{koromilas2026polysae,
  author={Koromilas, Panagiotis and Demou, Andreas D. and Oldfield, James and Panagakis, Yannis and Nicolaou, Mihalis A.},
  title={{PolySAE}: Modeling Feature Interactions in Sparse Autoencoders via Polynomial Decoding},
  year={2026},
  archivePrefix={arXiv},
  eprint={2602.01322},
  primaryClass={cs.LG},
  url={https://arxiv.org/abs/2602.01322},
  journal = {arXiv preprint arXiv:2602.01322}
}

@article{sunkumohan2026asb,
  title={Interpreting and Steering State-Space Models via Activation Subspace Bottlenecks},
  author={Sunku Mohan, Vamshi and Gupta, Kaustubh and Das, Aneesha and Singh, Chandan},
  year={2026},
  eprint={2602.22719},
  archivePrefix={arXiv},
  primaryClass={cs.LG},
  url={https://arxiv.org/abs/2602.22719},
  journal = {arXiv preprint arXiv:2602.22719}
}

@article{yap2026steeringmoe,
  title={Behavioral Steering in a 35{B} {M}o{E} Language Model via {SAE}-Decoded Probe Vectors: One Agency Axis, Not Five Traits},
  author={Yap, Jia Qing},
  year={2026},
  eprint={2603.16335},
  archivePrefix={arXiv},
  primaryClass={cs.LG},
  url={https://arxiv.org/abs/2603.16335},
  journal = {arXiv preprint arXiv:2603.16335}
}

@inproceedings{schlag2021fwp,
  title={Linear Transformers Are Secretly Fast Weight Programmers},
  author={Schlag, Imanol and Irie, Kazuki and Schmidhuber, J{\"u}rgen},
  booktitle={International Conference on Machine Learning (ICML)},
  year={2021},
  eprint        = {2102.11174},
  archivePrefix = {arXiv},
  primaryClass  = {cs.LG}
}

@inproceedings{lahoti2026mamba3,
  title={{Mamba}-3: Improved Sequence Modeling using State Space Principles},
  author={Lahoti, Aakash and Li, Kevin Y. and Chen, Berlin and Wang, Caitlin and Bick, Aviv and Kolter, J. Zico and Dao, Tri and Gu, Albert},
  booktitle={International Conference on Learning Representations (ICLR)},
  year={2026},
  eprint        = {2603.15569},
  archivePrefix = {arXiv},
  primaryClass  = {cs.LG}
}

@inproceedings{he2025lorsa,
  title={Towards Understanding the Nature of Attention with Low-Rank Sparse Decomposition},
  author={He, Zhengfu and Wang, Junxuan and Lin, Rui and Ge, Xuyang and Shu, Wentao and Tang, Qiong and Zhang, Junping and Qiu, Xipeng},
  booktitle={International Conference on Learning Representations (ICLR)},
  year={2026},
  eprint        = {2504.20938},
  archivePrefix = {arXiv},
  primaryClass  = {cs.LG}
}

@article{dao2024mamba2,
  title={Transformers are {SSMs}: Generalized Models and Efficient Algorithms through Structured State Space Duality},
  author={Dao, Tri and Gu, Albert},
  year={2024},
  eprint={2405.21060},
  archivePrefix={arXiv},
  primaryClass={cs.LG},
  url={https://arxiv.org/abs/2405.21060},
  booktitle={International Conference on Machine Learning (ICML)},
  journal = {arXiv preprint arXiv:2405.21060}
}

@inproceedings{yang2024gdn,
  title={Gated Delta Networks: Improving {Mamba2} with Delta Rule},
  author={Yang, Songlin and Kautz, Jan and Hatamizadeh, Ali},
  booktitle={International Conference on Learning Representations (ICLR)},
  year={2025},
  eprint        = {2412.06464},
  archivePrefix = {arXiv},
  primaryClass  = {cs.LG}
}

@article{yang2023gla,
  title={Gated Linear Attention Transformers with Hardware-Efficient Training},
  author={Yang, Songlin and Wang, Bailin and Shen, Yikang and Panda, Rameswar and Kim, Yoon},
  year={2024},
  eprint={2312.06635},
  archivePrefix={arXiv},
  primaryClass={cs.LG},
  url={https://arxiv.org/abs/2312.06635},
  booktitle={International Conference on Machine Learning (ICML)},
  journal = {arXiv preprint arXiv:2312.06635}
}

@inproceedings{yang2024deltanet,
  title={Parallelizing Linear Transformers with the Delta Rule over Sequence Length},
  author={Yang, Songlin and Wang, Bailin and Zhang, Yu and Shen, Yikang and Kim, Yoon},
  booktitle={Advances in Neural Information Processing Systems},
  pages={115491--115522},
  year={2024},
  doi={10.52202/079017-3668}
}

@inproceedings{peng2025rwkv7,
  title={{RWKV}-7 ``{Goose}'' with Expressive Dynamic State Evolution},
  author={Peng, Bo and Zhang, Ruichong and Goldstein, Daniel and Alcaide, Eric and Du, Xingjian and Hou, Haowen and Lin, Jiaju and Liu, Jiaxing and Lu, Janna and Merrill, William and Song, Guangyu and Tan, Kaifeng and Utpala, Saiteja and Wilce, Nathan and Wind, Johan S. and Wu, Tianyi and Wuttke, Daniel and Zhou-Zheng, Christian},
  booktitle={Conference on Language Modeling (COLM)},
  year={2025},
  eprint        = {2503.14456},
  archivePrefix = {arXiv},
  primaryClass  = {cs.LG}
}

@misc{hu2025comba,
  title={Comba: Improving Bilinear {RNN}s with Closed-loop Control},
  author={Hu, Jiaxi and Pan, Yongqi and Du, Jusen and Lan, Disen and Tang, Xiaqiang and Wen, Qingsong and Liang, Yuxuan and Sun, Weigao},
  year={2025},
  eprint={2506.02475},
  archivePrefix={arXiv},
  primaryClass={cs.LG},
  doi={10.48550/arxiv.2506.02475},
  url={https://arxiv.org/abs/2506.02475}
}

@article{lieberum2024gemmascope,
  title={Gemma Scope: Open Sparse Autoencoders Everywhere All At Once on {G}emma 2},
  author={Lieberum, Tom and Rajamanoharan, Senthooran and Conmy, Arthur and Smith, Lewis and Sonnerat, Nicolas and Varma, Vikrant and Kram{\'a}r, J{\'a}nos and Dragan, Anca and Shah, Rohin and Nanda, Neel},
  year={2024},
  eprint={2408.05147},
  archivePrefix={arXiv},
  primaryClass={cs.LG},
  url={https://arxiv.org/abs/2408.05147},
  booktitle={BlackboxNLP 2024},
  journal = {arXiv preprint arXiv:2408.05147}
}

@article{engels2024notlinear,
  title={Not All Language Model Features Are One-Dimensionally Linear},
  author={Engels, Joshua and Michaud, Eric J. and Liao, Isaac and Gurnee, Wes and Tegmark, Max},
  year={2025},
  eprint={2405.14860},
  archivePrefix={arXiv},
  primaryClass={cs.LG},
  url={https://arxiv.org/abs/2405.14860},
  booktitle={International Conference on Learning Representations (ICLR)},
  journal = {arXiv preprint arXiv:2405.14860}
}

@article{dunefsky2024transcoders,
  title={Transcoders Find Interpretable {LLM} Feature Circuits},
  author={Dunefsky, Jacob and Chlenski, Philippe and Nanda, Neel},
  year={2024},
  eprint={2406.11944},
  archivePrefix={arXiv},
  primaryClass={cs.LG},
  url={https://arxiv.org/abs/2406.11944},
  booktitle={Advances in Neural Information Processing Systems (NeurIPS)},
  journal = {arXiv preprint arXiv:2406.11944}
}

@inproceedings{sharma2024mamba,
  title={Locating and Editing Factual Associations in {M}amba},
  author={Sharma, Arnab Sen and Atkinson, David and Bau, David},
  booktitle={Conference on Language Modeling (COLM)},
  year={2024},
  eprint        = {2404.03646},
  archivePrefix = {arXiv},
  primaryClass  = {cs.LG}
}

@misc{axbench,
  title={{AxBench}: Steering {LLMs}? {E}ven Simple Baselines Outperform Sparse Autoencoders},
  author={Wu, Zhengxuan and Arora, Aryaman and Geiger, Atticus and Wang, Zheng and Huang, Jing and Jurafsky, Dan and Manning, Christopher D. and Potts, Christopher},
  year={2025},
  archivePrefix={arXiv},
  eprint={2501.17148},
  primaryClass={cs.CL},
  url={https://arxiv.org/abs/2501.17148},
  booktitle={International Conference on Machine Learning (ICML)}
}

@misc{matryoshkasae,
  title={Learning Multi-Level Features with Matryoshka Sparse Autoencoders},
  author={Bussmann, Bart and Nabeshima, Noa and Karvonen, Adam and Nanda, Neel},
  year={2025},
  archivePrefix={arXiv},
  eprint={2503.17547},
  primaryClass={cs.LG},
  url={https://arxiv.org/abs/2503.17547},
  booktitle={International Conference on Machine Learning (ICML)}
}

@misc{batchtopk,
  title={{BatchTopK} Sparse Autoencoders},
  author={Bussmann, Bart and Leask, Patrick and Nanda, Neel},
  year={2024},
  archivePrefix={arXiv},
  eprint={2412.06410},
  primaryClass={cs.LG},
  url={https://arxiv.org/abs/2412.06410}
}

@inproceedings{katharopoulos2020linear,
  title={Transformers are {RNN}s: Fast Autoregressive Transformers with Linear Attention},
  author={Katharopoulos, Angelos and Vyas, Apoorv and Pappas, Nikolaos and Fleuret, Fran{\c{c}}ois},
  booktitle={International Conference on Machine Learning (ICML)},
  pages={5156--5165},
  year={2020},
  eprint        = {2006.16236},
  archivePrefix = {arXiv},
  primaryClass  = {cs.LG}
}

@article{schmidhuber1992fwp,
  title={Learning to Control Fast-Weight Memories: An Alternative to Dynamic Recurrent Networks},
  author={Schmidhuber, J{\"u}rgen},
  journal={Neural Computation},
  volume={4},
  number={1},
  pages={131--139},
  year={1992},
  doi={10.1162/neco.1992.4.1.131}
}

@inproceedings{ba2016fastweights,
  title={Using Fast Weights to Attend to the Recent Past},
  author={Ba, Jimmy and Hinton, Geoffrey and Mnih, Volodymyr and Leibo, Joel Z. and Ionescu, Catalin},
  booktitle={Advances in Neural Information Processing Systems (NeurIPS)},
  year={2016},
  eprint        = {1610.06258},
  archivePrefix = {arXiv},
  primaryClass  = {cs.LG}
}

@article{olshausen1996sparse,
  title={Emergence of simple-cell receptive field properties by learning a sparse code for natural images},
  author={Olshausen, Bruno A. and Field, David J.},
  journal={Nature},
  volume={381},
  number={6583},
  pages={607--609},
  year={1996},
  doi={10.1038/381607a0}
}

@article{sun2024ttt,
  title={Learning to (Learn at Test Time): {RNN}s with Expressive Hidden States},
  author={Sun, Yu and Li, Xinhao and Dalal, Karan and Xu, Jiarui and Vikram, Arjun and Zhang, Genghan and Dubois, Yann and Chen, Xinlei and Wang, Xiaolong and Koyejo, Sanmi and Hashimoto, Tatsunori and Guestrin, Carlos},
  year={2024},
  archivePrefix={arXiv},
  eprint={2407.04620},
  primaryClass={cs.LG},
  url={https://arxiv.org/abs/2407.04620},
  journal = {arXiv preprint arXiv:2407.04620}
}

@misc{Gokaslan2019OpenWeb,
  author = {Gokaslan, Aaron and Cohen, Vanya},
  title = {OpenWebText Corpus},
  year = {2019},
  howpublished = {\url{https://skylion007.github.io/OpenWebTextCorpus}},
  url = {https://skylion007.github.io/OpenWebTextCorpus}}

@misc{qwen2025qwen3,
  title={{Qwen3} Technical Report},
  author={Yang, An and Li, Anfeng and Yang, Baosong and Zhang, Beichen and Hui, Binyuan and Zheng, Bo and Yu, Bowen and Gao, Chang and Huang, Chengen and Lv, Chenxu and Zheng, Chujie and Liu, Dayiheng and Zhou, Fan and Huang, Fei and Hu, Feng and Ge, Hao and Wei, Haoran and Lin, Huan and Tang, Jialong and Yang, Jian and Tu, Jianhong and Zhang, Jianwei and Yang, Jianxin and Yang, Jiaxi and Zhou, Jing and Zhou, Jingren and Lin, Junyang and Dang, Kai and Bao, Keqin and Yang, Kexin and Yu, Le and Deng, Lianghao and Li, Mei and Xue, Mingfeng and Li, Mingze and Zhang, Pei and Wang, Peng and Zhu, Qin and Men, Rui and Gao, Ruize and Liu, Shixuan and Luo, Shuang and Li, Tianhao and Tang, Tianyi and Yin, Wenbiao and Ren, Xingzhang and Wang, Xinyu and Zhang, Xinyu and Ren, Xuancheng and Fan, Yang and Su, Yang and Zhang, Yichang and Zhang, Yinger and Wan, Yu and Liu, Yuqiong and Wang, Zekun and Cui, Zeyu and Zhang, Zhenru and Zhou, Zhipeng and Qiu, Zihan},
  year={2025},
  archivePrefix={arXiv},
  eprint={2505.09388},
  primaryClass={cs.CL},
  url={https://arxiv.org/abs/2505.09388}
}

@misc{yang2024fla,
  title={Flash Linear Attention},
  author={Yang, Songlin and Zhang, Yu},
  year={2024},
  howpublished={\url{https://github.com/sustcsonglin/flash-linear-attention}},
  url={https://github.com/sustcsonglin/flash-linear-attention}
}

@misc{nazari2026rank,
  title={The Key to State Reduction in Linear Attention: A Rank-based Perspective},
  author={Nazari, Philipp and Rusch, T. Konstantin},
  year={2026},
  archivePrefix={arXiv},
  eprint={2602.04852},
  primaryClass={cs.LG},
  url={https://arxiv.org/abs/2602.04852}
}

@misc{sun2025densefeat,
  author = {Sun, Xiaoqing and Stolfo, Alessandro and Engels, Joshua and Wu, Ben and Rajamanoharan, Senthooran and Sachan, Mrinmaya and Tegmark, Max},
  title = {Dense {SAE} Latents Are Features, Not Bugs},
  year = {2025},
  archivePrefix = {arXiv},
  eprint = {2506.15679},
  primaryClass = {cs.LG},
  url = {https://arxiv.org/abs/2506.15679},
  booktitle={Advances in Neural Information Processing Systems (NeurIPS)}
}

@misc{paulo2025seedfeat,
  author = {Paulo, Gon{\c{c}}alo and Belrose, Nora},
  title = {Sparse Autoencoders Trained on the Same Data Learn Different Features},
  year = {2026},
  archivePrefix = {arXiv},
  eprint = {2501.16615},
  primaryClass = {cs.LG},
  url = {https://arxiv.org/abs/2501.16615},
  booktitle={International Conference on Learning Representations (ICLR)}
}

@misc{jiralerspong2026crossarch,
  author = {Jiralerspong, Thomas and Bricken, Trenton},
  title = {Cross-Architecture Model Diffing with Crosscoders: Unsupervised Discovery of Differences Between {LLMs}},
  year = {2026},
  archivePrefix = {arXiv},
  eprint = {2602.11729},
  primaryClass = {cs.LG},
  url = {https://arxiv.org/abs/2602.11729}
}

@misc{lan2024univsae,
  author = {Lan, Michael and Torr, Philip and Meek, Austin and Khakzar, Ashkan and Krueger, David and Barez, Fazl},
  title = {Quantifying Feature Space Universality Across Large Language Models via Sparse Autoencoders},
  year = {2024},
  archivePrefix = {arXiv},
  eprint = {2410.06981},
  primaryClass = {cs.LG},
  url = {https://arxiv.org/abs/2410.06981}
}

@misc{bussmann2025matchpursuit,
  author = {Costa, Val{\'e}rie and Fel, Thomas and Lubana, Ekdeep Singh and Tolooshams, Bahareh and Ba, Demba},
  title = {Evaluating Sparse Autoencoders: From Shallow Design to Matching Pursuit},
  year = {2025},
  archivePrefix = {arXiv},
  eprint = {2506.05239},
  primaryClass = {cs.LG},
  url = {https://arxiv.org/abs/2506.05239}
}

@misc{paulo2025skiptranscoder,
  title={Transcoders Beat Sparse Autoencoders for Interpretability},
  author={Paulo, Gon{\c c}alo and Shabalin, Stepan and Belrose, Nora},
  year={2025}, archivePrefix={arXiv}, primaryClass={cs.LG}, eprint={2501.18823},
  url={https://arxiv.org/abs/2501.18823}
}

@misc{elhage2022toymodels,
  title={Toy Models of Superposition},
  author={Elhage, Nelson and Hume, Tristan and Olsson, Catherine and Schiefer, Nicholas and Henighan, Tom and Kravec, Shauna and Hatfield-Dodds, Zac and Lasenby, Robert and Drain, Dawn and Chen, Carol and Grosse, Roger and McCandlish, Sam and Kaplan, Jared and Amodei, Dario and Wattenberg, Martin and Olah, Christopher},
  year={2022}, archivePrefix={arXiv}, primaryClass={cs.LG}, eprint={2209.10652},
  url={https://arxiv.org/abs/2209.10652}
}

@inproceedings{wu2024reft,
  title={{ReFT}: Representation Finetuning for Language Models},
  author={Wu, Zhengxuan and Arora, Aryaman and Wang, Zheng and Geiger, Atticus and Jurafsky, Dan and Manning, Christopher D. and Potts, Christopher},
  booktitle={Advances in Neural Information Processing Systems (NeurIPS)},
  year={2024},
  eprint        = {2404.03592},
  archivePrefix = {arXiv},
  primaryClass  = {cs.LG}
}

@misc{galim2024peftssm,
  title={Parameter-Efficient Fine-Tuning of State Space Models},
  author={Galim, Kevin and Kang, Wonjun and Zeng, Yuchen and Koo, Hyung Il and Lee, Kangwook},
  year={2024}, archivePrefix={arXiv}, primaryClass={cs.LG}, eprint={2410.09016},
  url={https://arxiv.org/abs/2410.09016}
}

@misc{makhzani2013ksparse,
  title={k-Sparse Autoencoders},
  author={Makhzani, Alireza and Frey, Brendan},
  year={2013}, archivePrefix={arXiv}, primaryClass={cs.LG}, eprint={1312.5663},
  url={https://arxiv.org/abs/1312.5663}
}

@misc{du2025mom,
  title={{MoM}: Linear Sequence Modeling with Mixture-of-Memories},
  author={Du, Jusen and Sun, Weigao and Lan, Disen and Hu, Jiaxi and Cheng, Yu},
  year={2026}, archivePrefix={arXiv}, primaryClass={cs.LG}, eprint={2502.13685},
  url={https://arxiv.org/abs/2502.13685},
  booktitle={International Conference on Learning Representations (ICLR)}
}

@inproceedings{meng2023memit,
  title={Mass-Editing Memory in a Transformer},
  author={Meng, Kevin and Sharma, Arnab Sen and Andonian, Alex and Belinkov, Yonatan and Bau, David},
  booktitle={International Conference on Learning Representations (ICLR)},
  year={2023},
  eprint        = {2210.07229},
  archivePrefix = {arXiv},
  primaryClass  = {cs.LG}
}

@misc{sun2023retnet,
  title={Retentive Network: A Successor to Transformer for Large Language Models},
  author={Sun, Yutao and Dong, Li and Huang, Shaohan and Ma, Shuming and Xia, Yuqing and Xue, Jilong and Wang, Jianyong and Wei, Furu},
  year={2023}, archivePrefix={arXiv}, primaryClass={cs.LG}, eprint={2307.08621},
  url={https://arxiv.org/abs/2307.08621}
}

@misc{scherlis2022polysemanticity,
  title={Polysemanticity and Capacity in Neural Networks},
  author={Scherlis, Adam and Sachan, Kshitij and Jermyn, Adam S. and Benton, Joe and Shlegeris, Buck},
  year={2022}, archivePrefix={arXiv}, primaryClass={cs.LG}, eprint={2210.01892},
  url={https://arxiv.org/abs/2210.01892}
}

@misc{gurnee2023sparseprobing,
  title={Finding Neurons in a Haystack: Case Studies with Sparse Probing},
  author={Gurnee, Wes and Nanda, Neel and Pauly, Matthew and Harvey, Katherine and Troitskii, Dmitrii and Bertsimas, Dimitris},
  year={2023}, archivePrefix={arXiv}, primaryClass={cs.LG}, eprint={2305.01610},
  url={https://arxiv.org/abs/2305.01610}
}

@inproceedings{zhang2024hedgehog,
  title={The Hedgehog \& the Porcupine: Expressive Linear Attentions with Softmax Mimicry},
  author={Zhang, Michael and Bhatia, Kush and Kumbong, Hermann and R{\'e}, Christopher},
  booktitle={International Conference on Learning Representations (ICLR)},
  year={2024},
  eprint        = {2402.04347},
  archivePrefix = {arXiv},
  primaryClass  = {cs.LG}
}

@misc{farnik2025jacobian,
  title={Jacobian Sparse Autoencoders: Sparsify Computations, Not Just Activations},
  author={Farnik, Lucy and Lawson, Tim and Houghton, Conor and Aitchison, Laurence},
  year={2025}, archivePrefix={arXiv}, primaryClass={cs.LG}, eprint={2502.18147},
  url={https://arxiv.org/abs/2502.18147},
  booktitle={International Conference on Machine Learning (ICML)}
}

@misc{braun2025apd,
  title={Interpretability in Parameter Space: Minimizing Mechanistic Description Length with Attribution-based Parameter Decomposition},
  author={Braun, Dan and Bushnaq, Lucius and Heimersheim, Stefan and Mendel, Jake and Sharkey, Lee},
  year={2025}, archivePrefix={arXiv}, primaryClass={cs.LG}, eprint={2501.14926},
  url={https://arxiv.org/abs/2501.14926}
}

@misc{bushnaq2025spd,
  title={Stochastic Parameter Decomposition},
  author={Bushnaq, Lucius and Braun, Dan and Sharkey, Lee},
  year={2025}, archivePrefix={arXiv}, primaryClass={cs.LG}, eprint={2506.20790},
  url={https://arxiv.org/abs/2506.20790}
}

@misc{siems2025deltaproduct,
  title={DeltaProduct: Improving State-Tracking in Linear RNNs via Householder Products},
  author={Siems, Julien and Carstensen, Timur and Zela, Arber and Hutter, Frank and Pontil, Massimiliano and Grazzi, Riccardo},
  year={2025}, archivePrefix={arXiv}, primaryClass={cs.LG}, eprint={2502.10297},
  url={https://arxiv.org/abs/2502.10297}
}

@misc{chang2025xkv,
  title={{xKV}: Cross-Layer {SVD} for {KV}-Cache Compression},
  author={Chang, Chi-Chih and Lin, Chien-Yu and Akhauri, Yash and Lin, Wei-Cheng and Wu, Kai-Chiang and Ceze, Luis and Abdelfattah, Mohamed S.},
  year={2025}, archivePrefix={arXiv}, eprint={2503.18893},
  primaryClass={cs.LG},
  url={https://arxiv.org/abs/2503.18893}
}

@misc{ravishankar2015soupdil,
  title={Efficient Sum of Outer Products Dictionary Learning ({SOUP-DIL}) - The $\ell_0$ Method},
  author={Ravishankar, Saiprasad and Nadakuditi, Raj Rao and Fessler, Jeffrey A.},
  year={2015}, archivePrefix={arXiv}, eprint={1511.08842},
  primaryClass={cs.LG},
  url={https://arxiv.org/abs/1511.08842}
}

@misc{pitorro2025latim,
  title={{LaTIM}: Measuring Latent Token-to-Token Interactions in {Mamba} Models},
  author={Pitorro, Hugo and Treviso, Marcos},
  year={2025}, archivePrefix={arXiv}, eprint={2502.15612},
  primaryClass={cs.LG},
  url={https://arxiv.org/abs/2502.15612}
}

@misc{airlangga2025mamba,
  title={Emergence of Primacy and Recency Effect in {Mamba}: A Mechanistic Point of View},
  author={Airlangga, Muhammad Cendekia and AlQuabeh, Hilal and Nwadike, Munachiso S. and Inui, Kentaro},
  year={2025}, archivePrefix={arXiv}, eprint={2506.15156},
  primaryClass={cs.LG},
  url={https://arxiv.org/abs/2506.15156}
}

@misc{arora2025ssmevals,
  title={Mechanistic Evaluation of Transformers and State Space Models},
  author={Arora, Aryaman and Rathi, Neil and Selvam, Nikil Roashan and Csord{\'a}s, R{\'o}bert and Jurafsky, Dan and Potts, Christopher},
  year={2025}, archivePrefix={arXiv}, eprint={2505.15105},
  primaryClass={cs.LG},
  url={https://arxiv.org/abs/2505.15105}
}

@misc{okpekpe2025recall,
  title={Revisiting Associative Recall in Modern Recurrent Models},
  author={Okpekpe, Destiny and Orvieto, Antonio},
  year={2025}, archivePrefix={arXiv}, eprint={2508.19029},
  primaryClass={cs.LG},
  url={https://arxiv.org/abs/2508.19029}
}

@misc{mueller2025mib,
  title={{MIB}: A Mechanistic Interpretability Benchmark},
  author={Mueller, Aaron and Geiger, Atticus and Wiegreffe, Sarah and Arad, Dana and Arcuschin, Iv{\'a}n and Belfki, Adam and Chan, Yik Siu and Fiotto-Kaufman, Jaden and Haklay, Tal and Hanna, Michael and Huang, Jing and Gupta, Rohan and Nikankin, Yaniv and Orgad, Hadas and Prakash, Nikhil and Reusch, Anja and Sankaranarayanan, Aruna and Shao, Shun and Stolfo, Alessandro and Tutek, Martin and Zur, Amir and Bau, David and Belinkov, Yonatan},
  year={2026}, archivePrefix={arXiv}, eprint={2504.13151},
  primaryClass={cs.LG},
  url={https://arxiv.org/abs/2504.13151},
  booktitle={International Conference on Machine Learning (ICML)}
}

@misc{zhang2023patching,
  title={Towards Best Practices of Activation Patching in Language Models: Metrics and Methods},
  author={Zhang, Fred and Nanda, Neel},
  year={2024}, archivePrefix={arXiv}, eprint={2309.16042},
  primaryClass={cs.LG},
  url={https://arxiv.org/abs/2309.16042},
  booktitle={International Conference on Learning Representations (ICLR)}
}

\newpage
\appendix

\section{Derivation of the Three-Factor Logit Factorization}
\label{app:closed_form}

\begin{figure}[!htbp]
  \centering
  \begin{minipage}[b]{0.56\linewidth}
    \centering
    \includegraphics[width=\linewidth]{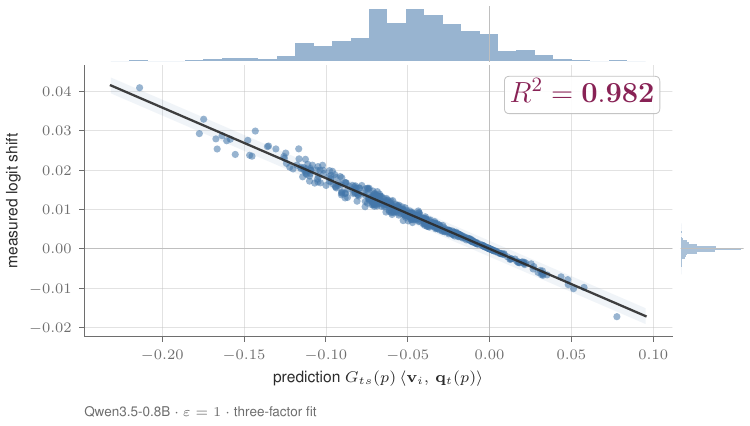}\\
    {\small (a) Single L9 H4 feature, $R^2{=}0.982$.}
  \end{minipage}\hfill
  \begin{minipage}[b]{0.42\linewidth}
    \centering
    \includegraphics[width=\linewidth]{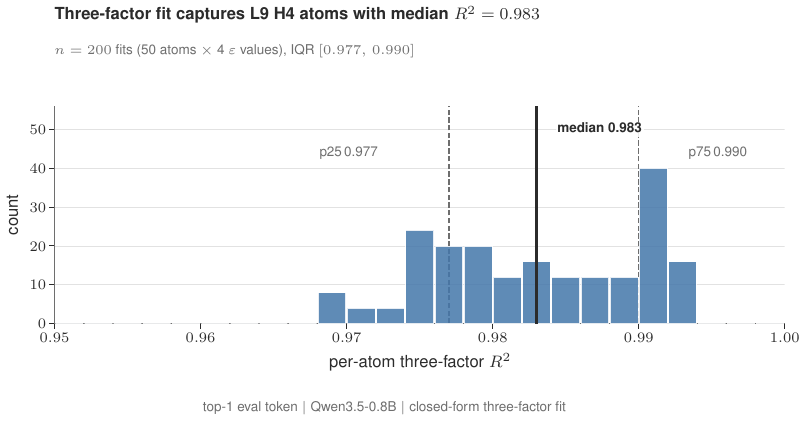}\\
    {\small (b) Population: median $R^2{=}0.983$, IQR $[0.977, 0.990]$.}
  \end{minipage}
  \caption{\textbf{Rank-1 state perturbations follow a three-factor logit expression.} (a) Measured logit shift vs.~predicted $G_{t_0 \to t}(c) \cdot \langle \mathbf{w}_i, \mathbf{q}_t(c) \rangle \cdot \langle \mathbf{v}_i, W_U[\mathrm{tok}] \rangle$ for one L9 H4 feature. (b) Per-atom three-factor $R^2$ across $n{=}200$ fits ($50$ atoms $\times$ $4$ $\varepsilon$).}
  \label{fig:closed_form}
\end{figure}

Under a rank-$1$ perturbation of the cached \gdn state at reference position $t_0 < t$ along feature $i$ with decoder pair $(\mathbf{v}_i, \mathbf{w}_i)$,
\begin{equation}
\label{eq:cfl}
\Delta\ell_{\mathrm{tok}}(c, i, t) \approx G_{t_0 \to t}(c) \cdot \langle \mathbf{w}_i, \mathbf{q}_t(c) \rangle \cdot \langle \mathbf{v}_i, W_U[\mathrm{tok}] \rangle,
\end{equation}
where $\mathbf{q}_t$ is the query at $t$, $W_U[\mathrm{tok}]$ the output-embedding row, and $G_{t_0\to t}$ a prompt-/position-specific gate product.

\paragraph{Setup.}
Let $S_{t_0}$ be the cached \gdn state at reference position $t_0$ for one head at one layer. A rank-1 perturbation along feature $i$'s decoder writes $S_{t_0} \mapsto S_{t_0} + \varepsilon\, \mathbf{v}_i \mathbf{w}_i^\top$. We propagate through the remaining \gdn recurrence to $t$, read against $\mathbf{q}_t$, pass through residual, attention, and MLP layers, and project to logits via $W_U$.

\paragraph{State propagation.}
The native gated delta rule \citep{yang2024gdn} is
\[
S_{s+1} = \alpha_{s+1}(c)\bigl(I - \beta_{s+1}(c)\,\mathbf{k}_{s+1}(c)\mathbf{k}_{s+1}(c)^\top\bigr) S_s + \beta_{s+1}(c) \, \mathbf{k}_{s+1}(c) \mathbf{v}_{s+1}(c)^\top.
\]
Differencing perturbed and native trajectories cancels the additive write and leaves
\[
\delta S_{s+1} = \alpha_{s+1}(c)\bigl(I - \beta_{s+1}(c)\,\mathbf{k}_{s+1}(c)\mathbf{k}_{s+1}(c)^\top\bigr)\, \delta S_s, \qquad \delta S_{t_0} = \varepsilon\, \mathbf{v}_i \mathbf{w}_i^\top.
\]
Multiplying by the rank-one correction gives a cross term of magnitude $\beta_{s+1}\langle\mathbf{w}_i,\mathbf{k}_{s+1}\rangle$. For a register atom, $\mathbf{w}_i$ aligns with one prompt-restricted subset of keys; on every other step the cross term is small, so the correction acts like the identity on $\delta S_s$ and leaves the scalar gate $\alpha_{s+1}(c)$. Iterating,
\[
\delta S_t \;\approx\; \varepsilon\, \left[\prod_{s=t_0+1}^{t} \alpha_s(c)\right] \mathbf{v}_i \mathbf{w}_i^\top \;\equiv\; \varepsilon\, G^{\alpha}_{t_0 \to t}(c)\, \mathbf{v}_i \mathbf{w}_i^\top.
\]
The reduction is empirical: residual cross-term mass is accounted for by the same prompt-dependent prefactor we absorb into $G_{t_0\to t}(c)$. A scalar gate cannot mix the rank-1 outer with anything else, so the perturbation stays rank-1; only its norm decays.

\paragraph{Read through the head.}
At position $t$ the head reads $\mathbf{o}_t = S_t \mathbf{q}_t$. Hitting that read with $\delta S_t$:
\[
\delta\mathbf{o}_t = \varepsilon\, G^{\alpha}_{t_0 \to t}(c) \cdot \langle \mathbf{w}_i, \mathbf{q}_t(c) \rangle \cdot \mathbf{v}_i.
\]
The output-space perturbation is pinned in direction to $\mathbf{v}_i$ regardless of prompt; the prompt only sets magnitude.

\paragraph{Downstream path to logits.}
Write $J(c, t)$ for the Jacobian from head output $(L, t)$ to the final residual stream at $t$. To first order in $\varepsilon$,
\[
\Delta\ell_{\mathrm{tok}}(c, i, t) = \varepsilon\, G^{\alpha}_{t_0 \to t}(c) \cdot \langle \mathbf{w}_i, \mathbf{q}_t(c) \rangle \cdot \langle J(c, t) \mathbf{v}_i,\, W_U[\mathrm{tok}] \rangle.
\]
Eq.~\ref{eq:cfl} substitutes $\mathbf{v}_i$ for $J(c, t) \mathbf{v}_i$, absorbing the rotation-and-rescaling into $G$. Two empirical observations support the substitution: encoder/decoder cosine for register atoms is $0.94 \pm 0.02$ on the DeltaNet probe ($0.65 \pm 0.10$ at Qwen L9 H4), and we observe $0$ sign flips of $\Delta\ell$ relative to $\mathrm{sign}\langle \mathbf{w}_i, \mathbf{q}_t \rangle$ across $10{,}000$ trials (Wilson upper bound $0.04\%$). A prompt-dependent rotation off the unembed direction would flip signs.

\paragraph{Empirical $G$.}
$G$ depends on activation statistics through $J(c, t)$, fit numerically: one scalar per (prompt, feature, eval-token) triple, recovered by least squares. The fitted $G$ tracks $\prod \alpha_s$ to first order with a small prompt-level residual. Median per-feature $R^2{=}0.98$ at L9 H4 (Figure~\ref{fig:closed_form}).

\paragraph{Gate factor in Qwen3.5 \gdn.}
The Qwen3.5 \gdn block computes $g_s = -\exp(A^{\log}_h)\cdot\mathrm{softplus}(a_s(c) + b^{\mathrm{dt}}_h)$; forget gate $\alpha_s = \exp(g_s)$, write gate $\beta_s = \sigma(b_s)$, both scalar per head. For horizon $h$, $G^{\alpha}_{t_0 \to t_0 + h}(c) = \prod_{s = t_0+1}^{t_0+h} \exp(g_s(c))$. The full $G$ multiplies this by $\|J(c, t) \mathbf{v}_i\| / \|\mathbf{v}_i\|$ and the cosine between $\mathbf{u}_i$ and $\mathbf{v}_i$; none depends on the eval token, so $G$ enters as a single scalar.

\paragraph{Population audit.}
Over $50\times 4$ (atom, $\varepsilon$) cells with $\varepsilon\in\{0.1, 0.3, 1.0, 3.0\}$, $500$ prompts each: median $R^2{=}0.983$, IQR $[0.977, 0.990]$, $10$th percentile $0.974$; all $200$ cells exceed $R^2{=}0.95$. Top-$2$--top-$5$ eval-token ranks hold $R^2\in[0.983, 0.984]$; tail logits degrade to $R^2{\approx}0.05$ at rank $50$. The expression fits dominant logit shifts; tail contributions are higher-order Taylor terms.

\paragraph{Scope.}
The approximation is a first-order Taylor expansion around $\varepsilon = 0$, supported by selectivity being $\varepsilon$-invariant to four decimals across $\varepsilon \in [0.1, 3]$ (\Sref{sec:mechanism}). Host-architecture analogs at Mamba-2 L24 H0 and Qwen3.5-4B L12 H8 yield negative $R^2$, identifying $G$ as the architecture-specific component (\Sref{sec:falsification}).

\section{SAE-Family Invariance of the Partition}
\label{app:sae_variants}

The partition in \Sref{sec:mechanism} uses BatchTopK \citep{batchtopk}. A JumpReLU bilinear SAE \citep{rajamanoharan2024jumprelu} retrained on the same Qwen3.5-0.8B L9 H4 GDN state (matched optimizer/LR/$20$-epoch budget; $\lambda_\mathrm{sparsity}{=}10^{-5}$, $\theta_0{=}10^{-4}$, bandwidth $10^{-3}\to10^{-5}$ cosine; converges to Val MSE $5.71\times 10^{-6}$, $L_0{=}1{,}142$, zero dead) reproduces the partition.

\begin{table}[h]
\centering
\small
\caption{\textbf{Partition is stable under a JumpReLU sparsity swap and reaches $105\times$ within-SAE separation against BatchTopK's $29\times$.} Both objectives recover the same register class on the same GDN state; JumpReLU's adaptive threshold uses every feature. Qwen3.5-0.8B L9 H4; $\nf{=}2048$; $20$ epochs.}
\label{tab:sae_variants}
\begin{tabular}{lrrrrr}
\toprule
SAE objective & $n_\mathrm{reg}\,/\,n_\mathrm{bun}$ & dead & register $\cos$ & bundle $\cos$ & reg/bun \\
\midrule
BatchTopK ($k{=}32$, $L_0{=}32$)           & 222\,/\,94    & 1{,}732 & 0.262 & 0.009  & 29$\times$  \\
JumpReLU ($L_0 {\approx} 1{,}142$)         & 1{,}259\,/\,773 & 16    & 0.189 & 0.0018 & 105$\times$ \\
\bottomrule
\end{tabular}
\end{table}

\begin{figure}[!htbp]
  \centering
  \includegraphics[width=0.75\linewidth]{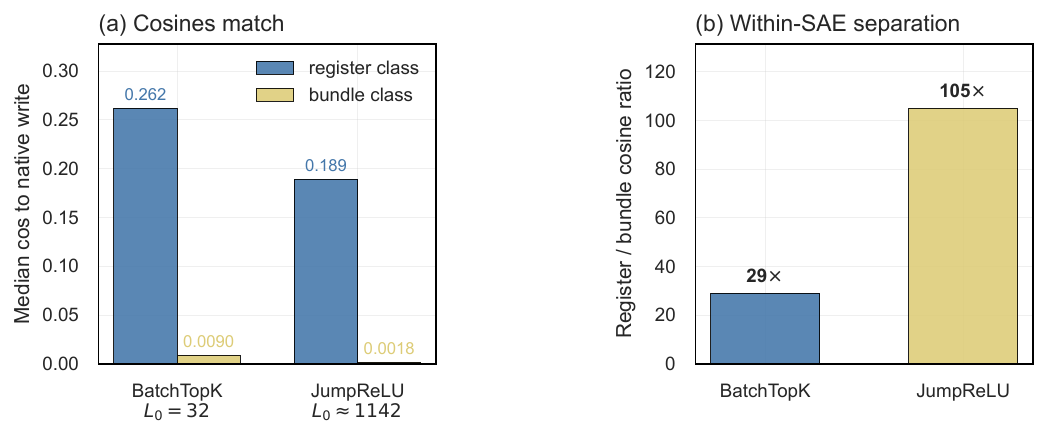}
  \caption{\textbf{Register/bundle partition is invariant to the sparsity mechanism.} (a) Median cosine to the native write under BatchTopK ($L_0{=}32$) and JumpReLU ($L_0 \approx 1{,}142$). Register cosines stay within $28\%$; bundle cosines are near zero in both. (b) Within-SAE register/bundle cosine ratio: JumpReLU $105\times$ vs BatchTopK $29\times$.}
  \label{fig:sae_family}
\end{figure}

\paragraph{Gated SAE (negative).}
Gated \citep{rajamanoharan2024gated} under hard, hard+STE, and soft-sigmoid ($\tau{=}0.1$) gate indicators all failed: hard converged to $L_0\to 0$ in epoch 1, STE stalled at $L_0{\approx}0.5$, and soft-sigmoid gave $7\times$ worse MSE at matched $L_0$. BatchTopK and JumpReLU are the two stable objectives.

\begin{table}[h]
\centering
\small
\caption{\textbf{The architecture-matched decoder gives the largest measured substitution-success gain where the native write is least rank-1.} \writesae{} register cosine measures the trained-dictionary side of the rank gradient (\Sref{sec:archscope}). Substitution-success gap is the percent of firings where the \writesae{} atom beats deleting the write minus the corresponding \flatsae{}+top-$K$-SVD rate on the same firing set. $\nf{=}2{,}048$, $k{=}32$, seed 42, 20 epochs; matched-Frobenius substitution; protocol in App.~\ref{app:flat_svd}.}
\label{tab:flat_svd_archprior}
\begin{tabular}{lrrr}
\toprule
Substrate (write rule) & \writesae{} reg.~$\cos$ & \writesae{} vs \flatsae{}+SVD $\Delta$ & $n_\mathrm{records}$ \\
\midrule
\gdn{} (rank-1 outer)         & $0.262$  & $-0.11$pp ($91.25\%$ vs $91.36\%$)\textsuperscript{$\dagger$} & $1{,}851$ \\
RWKV-7 (rank-2 outer + erase) & $0.180$  & $-2.52$pp ($45.3\%$ vs $47.8\%$)\textsuperscript{$\ast$} & $6{,}591$ \\
Mamba-2 (diagonal SSM)        & $0.0575$ & $+6.55$pp ($82.85\%$ vs $76.30\%$) & $2{,}367$ \\
\bottomrule
\end{tabular}
\\[2pt]
\footnotesize $^{\dagger}$\gdn{} gate decay already rank-1 dominates the state, so SVD top-1 of a flat-SAE atom recovers the same direction the trained dictionary picks; the two methods tie. $^{\ast}$Both RWKV-7 substitutions are at the deletion floor on this cell ($45.3\%$ and $47.8\%$ against a $50\%$ coin-flip baseline).
\end{table}

\paragraph{Mamba-2 random rank-1 baseline.}
For Mamba-2's diagonal-SSM substitution, the matched random rank-1 control samples $(w_\mathrm{head}, w_\mathrm{state})$ independently from the empirical per-coordinate distribution of native updates at the firing position, then renormalizes the resulting outer product to match the Frobenius norm of the native diagonal write. The single-cell number ($82.85\%$, $n{=}2{,}367$) is superseded by the $100$-atom population sweep at $88.08\%$ in \Sref{sec:archscope}.

\section{Full Encoder-Swap MSE Comparison}
\label{app:encoderswap}

\begin{table}[h]
\centering
\footnotesize
\caption{\textbf{Rank-1 reconstruction is within $8\%$ of the unconstrained flat upper bound at L1/L9 and $1.75\times$ worse at the diffuse L17.} Layer-averaged validation MSE ($\times 10^{-5}$, $16$ heads, $3$ seeds) across spectra spanning $\sigma_1/\sigma_2 \approx 12$ to $3.0$. Bold: lowest per layer. Flat-dense atoms in $\mathbb{R}^{128\times128}$ span the full state and mix many native writes per feature. Qwen3.5-0.8B; $\nf{=}2048$; $k{=}32$; $20$ epochs.}
\label{tab:encoder_swap_mse}
\begin{tabular}{lccc}
\toprule
& Layer~1 & Layer~9 & Layer~17 \\
& ($\sigma_1/\sigma_2 \approx 12$) & ($6.8$) & ($3.0$) \\
\midrule
Flat (dense enc., dense dec.)               & \textbf{2.61} & \textbf{2.58} & \textbf{30.1} \\
Rank-1 (dense enc., rank-1 dec.)            & 2.83 & 2.69 & 52.8 \\
Bilinear (bilinear enc., rank-1 dec.)       & 4.07 & 3.09 & 54.0 \\
Bilinear-flat (bilinear enc., dense dec.)   & 3.93 & 3.35 & 39.0 \\
Tied bilinear (bilinear enc., tied rank-1)  & 7.63 & 5.98 & 80.5 \\
\bottomrule
\end{tabular}
\end{table}

\paragraph{Rank-1 as architectural match.}
A \gdn write adds exactly one rank-1 outer $\mathbf{k}_t\mathbf{v}_t^\top$ per token; a rank-1 dictionary atom $\mathbf{v}_i\mathbf{w}_i^\top$ corresponds one-to-one with one cache event, and the substitution test of \Sref{sec:mechanism} patches exactly that event. A rank-$r$ atom patches $r$ writes per firing; a flat dense atom mixes up to $128$ writes per feature. The flat upper bound beats rank-1 by only $8\%$ at L1/L9 despite using $60\times$ more decoder parameters per atom, and by $1.75\times$ at the diffuse L17. The rank-1 prior trades reconstruction MSE for resolution of the write primitive.

\smallskip\noindent\textit{Parameter formulas} ($d_k{=}d_v{=}128$, $d_\mathrm{in}{=}d_k d_v{=}16{,}384$): \textbf{FlatSAE} $= 2\nf d_\mathrm{in} + \nf + d_\mathrm{in}$; \textbf{WriteSAE} $= \nf d_\mathrm{in} + \nf(d_k + d_v) + \nf + d_k d_v$; \textbf{BilinearSAE} $= 4\nf d_k + \nf + d_k d_v$.

\section{Hyperparameters}
\label{app:hyperparams}

Defaults at \texttt{core/train.py}; per-experiment scripts override only the deviations. All runs: Adam; MSE reconstruction with optional auxiliary dead-feature term ($\lambda_\mathrm{aux}{=}10^{-2}$, $k_\mathrm{aux}{=}256$); decoder column re-norm every $100$ steps; resampling every $250$ steps for atoms inactive $\geq 100$ steps; $80/20$ split with split seed $=$ weight-init seed; FP32 SAE params on BF16-cast activations.

\begin{table}[h]
\centering
\footnotesize
\caption{\textbf{Per-architecture training configuration.} Shared defaults (Adam, $80/20$ split, decoder re-norm every $100$ steps, resample every $250$ steps after $100$-step inactivity) apply throughout.}
\label{tab:hyperparams_arch}
\setlength{\tabcolsep}{4pt}
\resizebox{\textwidth}{!}{%
\begin{tabular}{@{}lccccc@{}}
\toprule
& \textbf{GDN (Qwen3.5-0.8B)} & \textbf{GDN (Qwen3.5-4B)} & \textbf{DeltaNet-1.3B} & \textbf{Mamba-2-370M} & \textbf{GLA-1.3B} \\
\midrule
Source script              & \texttt{sweeps/} & \texttt{run\_9pager\_overnight} & \texttt{run\_deltanet\_validation} & \texttt{mamba2/mamba2\_sae\_experiment} & \texttt{run\_gla\_validation} \\
Layers extracted           & 0,2,5,9,13,17,21 & matched (L12) & 1,12,22 & 0,6,14,31,46,47 & 1,12,22 \\
Heads per layer            & all (16) & all & head 0 & multi-head sweep & head 0 \\
Peak / Min LR              & $3{\times}10^{-4}$ / $3{\times}10^{-5}$ & same & same & same & same \\
Schedule                   & cosine + warmup & cosine + warmup & cosine + warmup & cosine + warmup & cosine + warmup \\
Warmup steps               & 50 & 50 & 50 & 100 & 50 \\
Batch / Epochs             & 256 / 20 & 256 / 20 & 256 / 20 & 128 / 50 & 256 / 20 \\
Sparsity / $k$             & TopK or BatchTopK / 32 & TopK / 32 & TopK / 32 & TopK / 32 & TopK / 32 \\
$n_\mathrm{features}$ / Decoder rank & $2048$ / 1 & $2048$ / 1 & $2048$ / 1 & $2048$ / 1 & $2048$ / 1 \\
Seeds                      & $\{0, 1, 42\}$ & $\{0, 1, 42\}$ & $\{0, 1, 2\}$ & $\{0, 1, 42\}$ & $\{0, 1, 2\}$ \\
Extraction (seq.\ len / \#) & $1024$ / up to $5{,}000$ & $1024$ / matched & $1024$ / $5{,}000$ & $1024$ / $5{,}000$ & $1024$ / $5{,}000$ \\
\bottomrule
\end{tabular}%
}
\end{table}

\begin{table}[h]
\centering
\footnotesize
\caption{\textbf{Per-experiment deviations from Table~\ref{tab:hyperparams_arch}.} Only listed fields differ from the matched-cohort GDN configuration.}
\label{tab:hyperparams_exp}
\setlength{\tabcolsep}{4pt}
\begin{tabular}{@{}p{0.30\linewidth}p{0.62\linewidth}@{}}
\toprule
\textbf{Experiment} & \textbf{Deviations from default} \\
\midrule
Encoder-swap ablation & SAE family $\in$ \{flat, rank1, bilinear, bilinear\_tied, bilinear\_flat\}; matched optimizer/LR/batch/epochs/$k$/$n_\mathrm{feat}$. \\
Higher-rank decoder sweep & rank $\in$ \{1, 2, 4\} on bilinear; otherwise default. \\
BatchTopK family-invariance & \texttt{use\_batchtopk=True}; same $k{=}32$. \\
JumpReLU family-invariance & sparsity rule $=$ JumpReLU, $\lambda_\mathrm{sparsity}{=}10^{-5}$, $\theta_0{=}10^{-4}$, bandwidth schedule cosine $10^{-3}\to10^{-5}$, $L_0{\approx}1142$. \\
Gated SAE (negative) & three gate variants tested; all collapsed or destabilized. \\
$k$-vs-head curve & $k$ swept $\in \{16, 32, 64\}$ at fixed $n_\mathrm{feat}{=}2048$. \\
Per-firing KL test & no SAE training; uses matched-cohort checkpoints. Cache deep-copied per conditional forward. \\
Rank-1 perturbation propagation (Experiment C) & no SAE training; closed-form prediction via $\hat\Delta = \mathbf{v}_i^\top S_t \mathbf{w}_i \cdot W_{\mathrm{out}}\mathrm{vec}(\mathbf{v}_i \mathbf{w}_i^\top)$, $\varepsilon \in \{0.25, 0.5, 0.75, 1.0\}$, $500$ prompts $\times$ $50$ features $\times$ $4$ scales $\times$ $64$ eval tokens, seed $2026$. \\
Steering / 4B held-out probe & inference-time only; matched-cohort 0.8B checkpoints with bilinear matched-filter encoder. \\
Mamba-2 multi-head sweep & 50 epochs, batch 128, warmup 100. \\
\bottomrule
\end{tabular}
\end{table}

\begin{table}[h]
\centering
\footnotesize
\caption{\textbf{\writesae{} architecture variants.} ``Bilinear'' encoder $a_i = \mathbf{v}_i^\top S_t \mathbf{w}_i$; ``Flat'' encoder is dense linear on $\mathrm{vec}(S_t)$. Dead-feature loss ($k_\mathrm{aux}{=}256$, $\lambda_\mathrm{aux}{=}10^{-2}$) and resampling cadence shared across rows.}
\label{tab:hyperparams_writesae}
\setlength{\tabcolsep}{4pt}
\begin{tabular}{@{}p{0.18\linewidth}p{0.21\linewidth}p{0.21\linewidth}p{0.06\linewidth}p{0.26\linewidth}@{}}
\toprule
\textbf{Variant} & \textbf{Encoder} & \textbf{Decoder} & \textbf{Bias} & \textbf{Norm constraint} \\
\midrule
\flatsae                   & dense linear on $\mathrm{vec}(S_t)$ & dense linear, $d_\mathrm{in}{\times}n_\mathrm{feat}$ & none & decoder column unit-norm \\
\ranksae                   & dense linear on $\mathrm{vec}(S_t)$ & rank-1 outer product $\mathbf{v}_i \mathbf{w}_i^\top$ & none & decoder factors unit-norm \\
\bilinsae                  & matched filter $\mathbf{v}_i^\top S_t \mathbf{w}_i$ & rank-1 outer product (untied factors) & none & encoder \& decoder factors unit-norm \\
\bilinsae\,(tied)          & matched filter (tied to decoder) & rank-1 outer product (factors shared with encoder) & none & shared factors unit-norm \\
\bilinsae\,(flat decoder)  & matched filter & dense linear, $d_\mathrm{in}{\times}n_\mathrm{feat}$ & none & decoder column unit-norm; encoder factors unit-norm \\
\bottomrule
\end{tabular}
\end{table}

\paragraph{Mamba-2 deviation.}
Per-head $d_\mathrm{state}{=}128$ produces a larger flat input than GDN; pilots had not converged at $20$ epochs, so we extend to $50$ epochs at batch $128$, $100$ warmup. Every other arch uses the GDN default block. All training scripts log the git \texttt{HEAD} SHA into \texttt{config.json}.

\paragraph{Training-budget control.}
At $5{\times}10^5$ states/head ($5$K corpus, $20$ epochs) the dead-feature ordering reverses to \ranksae{} $83.6\%$ $<$ \flatsae{} $86.6\%$ $<$ \bilinsae{} $93.3\%$; at $5{\times}10^6$ states/head with $200$ epochs the matched-cohort ordering returns (\bilinsae{} $85.8\%$ $<$ \flatsae{} $93.4\%$ $<$ \ranksae{} $94.0\%$). The $5$K corpus has covariance effective rank $70.2$ vs $74.7$ at $50$K and the same $\sigma_1/\sigma_2{=}5.998$, so the crossover is a training-dynamics effect, not a coverage gap. \bilinsae{}'s parameter-matched control in \Sref{sec:mechanism} ($5.5\times$ worse downstream PPL at the same $8.4$M budget) holds across schedules.

\section{Mechanism Support Figures}
\label{app:mech_figures}

\begin{figure}[!htbp]
  \centering
  \includegraphics[width=0.62\linewidth]{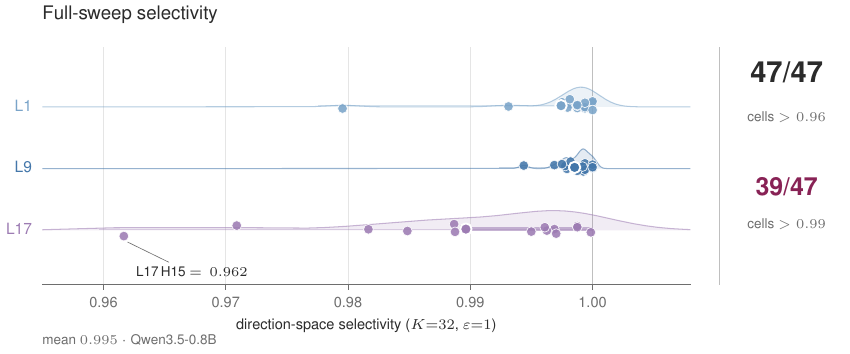}
  \caption{\textbf{Direction-space selectivity is high across the measured head sweep.} Each dot is one $(L, H)$ cell; horizontal position is per-cell mean selectivity, filled dot per-layer mean. Sweep $L\in\{1,9,17\}\times H\in\{0..15\}$ against matched-norm random rank-1 directions; L17 H14 excluded for upstream-cache corruption ($47/48$). Mean $0.9953$, $39/47$ cells exceed $0.99$. Qwen3.5-0.8B; $K{=}32$; $\varepsilon{=}1$.}
  \label{fig:selectivity}
\end{figure}

\begin{figure}[!htbp]
  \centering
  \includegraphics[width=0.62\linewidth]{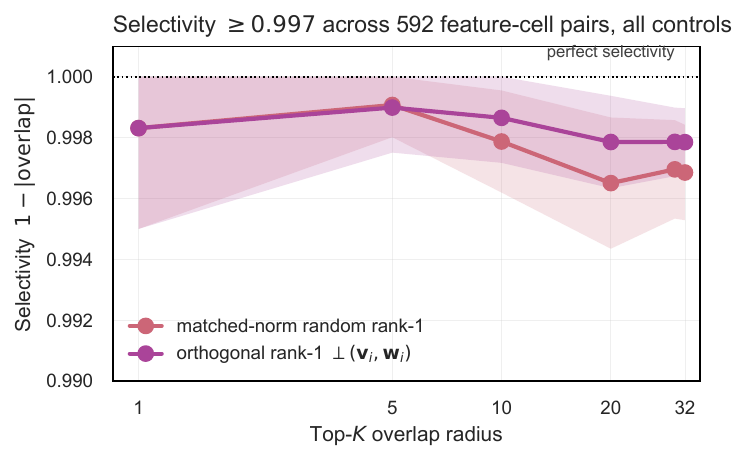}
  \caption{\textbf{Selectivity $\geq 0.997$ across 592 feature-cell pairs at every measured $K$ and every control.} Mean selectivity at Top-$K$ overlap $K\in\{1,5,10,20,30,32\}$ for matched-norm random rank-1 (red) and orthogonal rank-1 $\perp (\mathbf{v}_i,\mathbf{w}_i)$ (purple); flat-SVD coincides with random and is not drawn. Shaded bands $95\%$ CI over $n{=}592$ (layer, head, feature) triples; no control dips below $0.996$. Qwen3.5-0.8B L1/L9/L17.}
  \label{fig:selectivity_law}
\end{figure}

\begin{figure}[!htbp]
  \centering
  \includegraphics[width=0.95\linewidth]{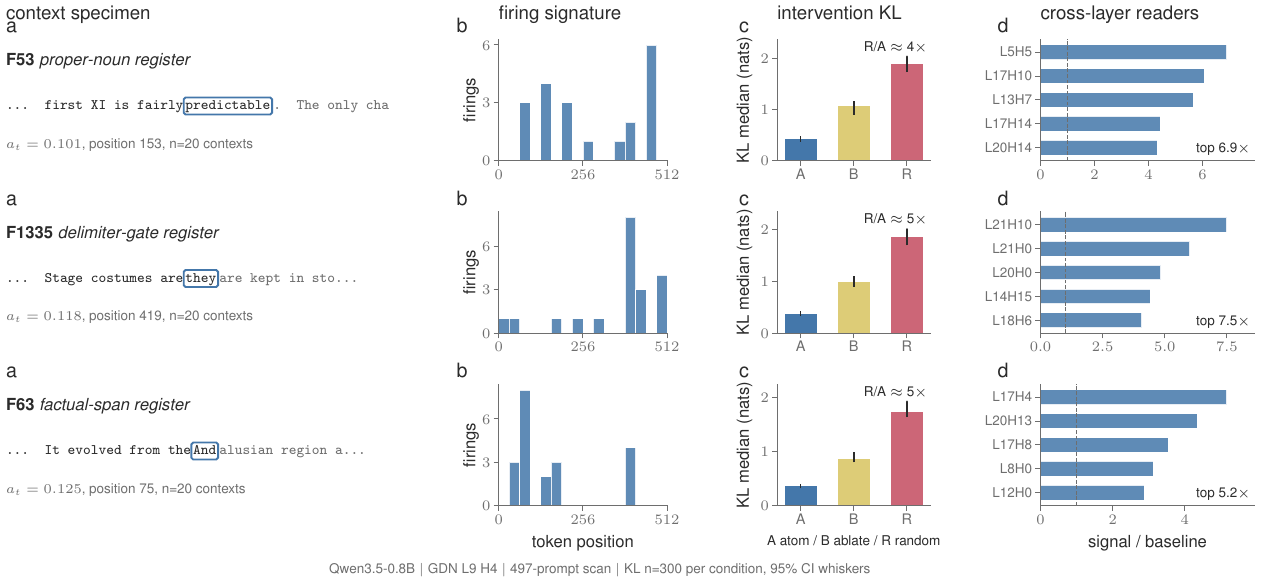}
  \caption{\textbf{Three register exemplars from Table~\ref{tab:exemplars} share write geometry and split surface roles.} Rows: F53 proper-noun register, F1335 delimiter gate, F63 factual-span register, all at Qwen3.5-0.8B GDN L9 H4. Columns: (a) one top-firing context with the firing token boxed, fire rate at left; (b) firing-rate histogram by token position over $497$ prompts, showing position-distributed firings rather than a positional artefact; (c) median KL at the final output distribution under matched-Frobenius rank-1 substitution (A: SAE atom, B: deleted write, R: random rank-1), with $95\%$ CI whiskers from $n{=}300$ per condition; (d) top cross-layer attention readers ranked by signal-to-baseline ratio. Atom substitutions remain near the deletion KL, while random controls increase KL by four to five times. Reader ratios in the $5.2$--$7.5\times$ range identify specific late-layer heads that read each atom's rank-1 write.}
  \label{fig:feature_gallery}
  \label{app:feature_gallery}
  \label{app:seedstability}
\end{figure}

\section{Cross-Architecture Partition and Scaling}
\label{app:scaling}

\begin{figure}[!htbp]
  \centering
  \includegraphics[width=\linewidth]{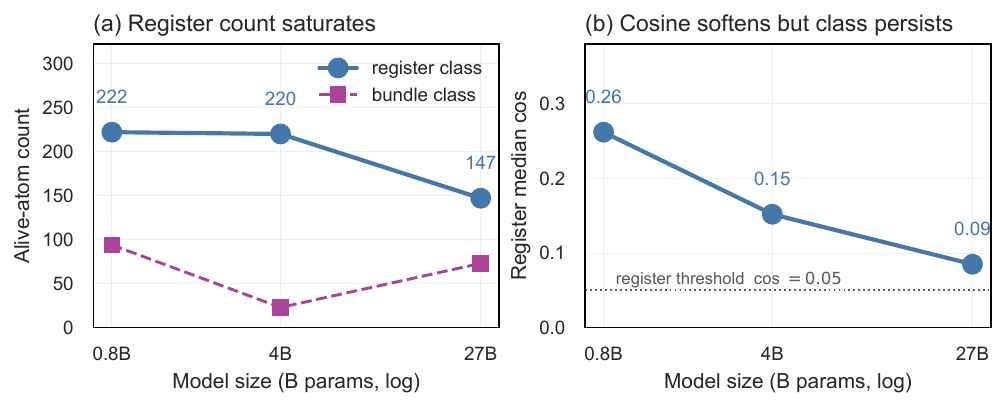}
  \caption{\textbf{Register class persists across the $34\times$ Qwen3.5 scale range.} (a) Alive-atom counts at 0.8B / 4B / 27B. Register count stable near $\sim\!220$ at 0.8B and 4B, $147$ at 27B. (b) Register median cosine softens from $0.26$ to $0.09$ but never crosses the register threshold $\cos{=}0.05$. Qwen3.5-0.8B L9 H4 / 4B L12 H8 / 27B L32 H16.}
  \label{fig:scale_saturation}
\end{figure}

\begin{table}[h]
  \centering
  \small
  \caption{\textbf{The $11.7\times$ DeltaNet-vs-Qwen27B gap (KS $p{=}1.2\!\times\!10^{-10}$) separates the tested host groups.} Outer-product-write cells have register/null ratios $60$--$383\times$ with high dead counts; diagonal or scalar-gated cells show lower ratios ($25$--$58\times$, dead $0$--$3$). DeltaNet L12 H8 $k{=}128$ reaches $\cos{=}0.997$ at $n_{\mathrm{reg}}{=}6$. Per-cell SAEs at $\nf{=}2048$.}
  \label{tab:cross_arch}
  \begin{tabular}{lrrrr}
    \toprule
    Configuration & $n_\mathrm{reg}\,/\,n_\mathrm{bun}$ & register median $\cos$ & ratio to null & dead \\
    \midrule
    \multicolumn{5}{l}{\emph{Outer-product write}} \\
    DeltaNet 1.3B L12 H8 $k{=}32$   & 0\,/\,118   & --              & --             & 1{,}930 \\
    DeltaNet 1.3B L12 H8 $k{=}64$   & 2\,/\,442   & 0.218           & 252$\times$    & 1{,}604 \\
    DeltaNet 1.3B L12 H8 $k{=}128$  & 6\,/\,425   & $\mathbf{0.997}$ & $\mathbf{383\times}$ & 1{,}617 \\
    DeltaNet 1.3B L6 H8 $k{=}64$    & 4\,/\,377   & 0.524           & 351$\times$    & 1{,}667 \\
    DeltaNet 1.3B L18 H8 $k{=}64$   & 21\,/\,110  & 0.209           & 85$\times$     & 1{,}917 \\
    Qwen3.5-0.8B L9 H4              & 222\,/\,94  & 0.262           & 192$\times$    & 1{,}732 \\
    Qwen3.5-4B L12 H8               & 220\,/\,23  & 0.152           & 116$\times$    & 1{,}805 \\
    Qwen3.5-27B L32 H16             & 147\,/\,73  & 0.085           & 60$\times$     & 1{,}828 \\
    \midrule
    \multicolumn{5}{l}{\emph{Diagonal or scalar-gated write}} \\
    Mamba-2 370M L24 H0 $k{=}64$    & 217\,/\,1{,}831 & 0.0575      & 58$\times$     & \textbf{0} \\
    GLA 1.3B L12 H0 $k{=}64$        & 564\,/\,1{,}481 & 0.110       & 25$\times$     & \textbf{3} \\
    \bottomrule
  \end{tabular}
\end{table}

\subsection{All L9 heads}
\label{app:all16}

Re-run of the firing-level test on every L9 head ($200$ firings per feature, $400$ prompts). The atom beats deletion on $89.29\% \pm 2.63\%$ of firings on average across $15$ heads with firings (range $82.61$--$93.20\%$); L9 H4 is $90.84\%$, $+0.59\sigma$ above the mean. $14/15$ heads exceed $85\%$, $12/15$ exceed $88\%$. The main-text cell is representative of L9, not selected on the dependent variable.

\begin{table}[h]
\centering
\small
\caption{\textbf{Per-head replacement results at Qwen3.5-0.8B L9.} L9 register feature set $[1442, 412, 192, 97, 1361, 53, 63, 87, 1335]$. H12 has no firings.}
\label{tab:all16_l9}
\begin{tabular}{lrccc}
\toprule
Head & $n_\mathrm{records}$ & atom wins \% & median $\mathrm{KL}_{\text{atom}}$ & median $\mathrm{KL}_{\text{delete}}$ \\
\midrule
H0  & $1{,}000$ & $91.90$ & $1.07$ & $1.90$ \\
H1  & $515$     & $89.13$ & $0.47$ & $1.04$ \\
H2  & $177$     & $85.31$ & $0.42$ & $0.96$ \\
H3  & $392$     & $88.27$ & $0.52$ & $1.08$ \\
\textbf{H4}  & $\mathbf{1{,}277}$ & $\mathbf{90.84}$ & $\mathbf{0.41}$ & $\mathbf{1.09}$ \\
H5  & $1{,}000$ & $93.20$ & $0.75$ & $1.56$ \\
H6  & $427$     & $88.52$ & $0.39$ & $0.88$ \\
H7  & $200$     & $92.50$ & $0.32$ & $0.78$ \\
H8  & $601$     & $87.35$ & $0.37$ & $0.87$ \\
H9  & $314$     & $89.81$ & $0.49$ & $1.11$ \\
H10 & $23$      & $82.61$ & $0.23$ & $0.54$ \\
H11 & $253$     & $90.12$ & $0.44$ & $0.95$ \\
H12 & --        & --      & --     & --     \\
H13 & $1{,}213$ & $89.37$ & $0.41$ & $1.01$ \\
H14 & $697$     & $90.10$ & $0.36$ & $0.84$ \\
H15 & $800$     & $90.38$ & $0.46$ & $1.17$ \\
\midrule
Mean (15 heads) & -- & $89.29 \pm 2.63$ & -- & -- \\
\bottomrule
\end{tabular}
\end{table}

\begin{figure}[h]
\centering
\includegraphics[width=0.55\textwidth]{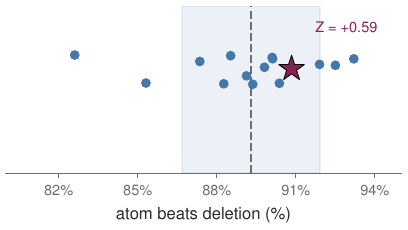}
\caption{\textbf{L9 H4 lies within the bulk of the per-head distribution.} Win rate across all $15$ L9 heads with firings (mean $89.29\% \pm 2.63\%$). Red star marks L9 H4 at $90.84\%$.}
\label{fig:all16_strip}
\end{figure}

\subsection{Population-level 4-way KL test at L9 H4}
\label{app:population_kl}

Same protocol as \Sref{sec:mechvalid} extended from the 8-feature example pool to the alive-atom population: $94$ bundle atoms ($\widetilde{\cos}{<}0.05$) plus $61$ stratified register atoms, $155$ total, capped at $30$ firings per atom. $87$ atoms reach the $\geq 5$-firing inclusion threshold.

Across $87$ atoms, the atom beats deletion on $\mathbf{89.80\%}$ of firings on average ($95\%$ CI $[88.1, 91.3]$ over $2{,}426$ firings; median $90.0\%$, range $60.9$--$100\%$). Class breakdown: $91.36\%$ register ($n{=}30$), $88.98\%$ bundle ($n{=}57$); Mann-Whitney $p{=}0.239$. Pearson $r{=}0.19$ between cosine and per-atom win rate ($p{=}0.08$); Spearman $\rho{=}0.06$ ($p{=}0.60$). Cosine to the native write tracks dictionary geometry; it does not predict substitution success at firing-level resolution.

\begin{table}[h]
\centering
\small
\caption{\textbf{The atom beats deletion on $89.80\%$ of firings across $87$ alive atoms ($95\%$ CI $[88.1, 91.3]$).} Win rates are near $\sim\!90\%$ across cosine bins except $[0.05, 0.20)$ where $10$ atoms straddle the threshold.}
\label{tab:population_kl_cosine_bins}
\setlength{\tabcolsep}{6pt}
\begin{tabular}{lrrr}
\toprule
Cosine bin & $n_\mathrm{atoms}$ & $n_\mathrm{firings}$ & atom wins \% \\
\midrule
$\cos < 0.00$           & $26$ & $481$   & $91.5$ \\
$0.00 \leq \cos < 0.05$ & $68$ & $1{,}069$ & $88.0$ \\
$0.05 \leq \cos < 0.20$ & $10$ & $122$   & $81.1$ \\
$0.20 \leq \cos < 0.30$ & $33$ & $480$   & $92.9$ \\
$\cos \geq 0.30$        & $18$ & $274$   & $92.7$ \\
\midrule
All atoms (firing-level)   & $155$ & $2{,}426$ & $89.85$ \\
$\geq 5$ firings (per-atom mean) & $87$ & $2{,}426$ & $\mathbf{89.80}$ \\
\bottomrule
\end{tabular}
\end{table}

\subsection{Per-head rank-1 vs rank-2 reconstruction at L9}
\label{app:rank2_perhead}

\begin{table}[h]
\centering
\small
\caption{\textbf{Per-head rank-1 vs rank-2 at Qwen3.5-0.8B L9.} Rank-2 lowers mean validation MSE by $3.1\%$ and wins on $11/15$ heads with both ranks trained, but the all-head substitution gives downstream perplexity $20.360$ at rank-2 vs $20.347$ at rank-1.}
\label{tab:rank2_perhead_l9}
\setlength{\tabcolsep}{4pt}
\begin{tabular}{lrrcrc}
\toprule
Head & $\mathrm{MSE}_{r_1}$ & $\mathrm{MSE}_{r_2}$ & $\mathrm{MSE}_{r_2}/\mathrm{MSE}_{r_1}$ & $n_\mathrm{records}$ & atom wins \% \\
\midrule
H0  & $5.05\!\times\!10^{-6}$  & $4.92\!\times\!10^{-6}$  & $0.974$ & $1{,}000$ & $91.90$ \\
H1  & $3.04\!\times\!10^{-6}$  & $3.09\!\times\!10^{-6}$  & $1.017$ & $515$     & $89.13$ \\
H2  & $6.76\!\times\!10^{-6}$  & $6.53\!\times\!10^{-6}$  & $0.966$ & $177$     & $85.31$ \\
H3  & $3.00\!\times\!10^{-6}$  & $2.96\!\times\!10^{-6}$  & $0.985$ & $392$     & $88.27$ \\
\textbf{H4}  & $\mathbf{2.17\!\times\!10^{-5}}$ & $\mathbf{2.17\!\times\!10^{-5}}$ & $\mathbf{1.000}$ & $\mathbf{1{,}277}$ & $\mathbf{90.84}$ \\
H5  & $1.01\!\times\!10^{-6}$  & $9.81\!\times\!10^{-7}$  & $0.972$ & $1{,}000$ & $93.20$ \\
H6  & $5.14\!\times\!10^{-6}$  & $5.10\!\times\!10^{-6}$  & $0.994$ & $427$     & $88.52$ \\
H7  & $1.38\!\times\!10^{-5}$  & $1.32\!\times\!10^{-5}$  & $0.950$ & $200$     & $92.50$ \\
H8  & $1.60\!\times\!10^{-5}$  & $1.57\!\times\!10^{-5}$  & $0.984$ & $601$     & $87.35$ \\
H9  & $6.84\!\times\!10^{-6}$  & $6.51\!\times\!10^{-6}$  & $0.952$ & $314$     & $89.81$ \\
H10 & $1.39\!\times\!10^{-5}$  & $1.21\!\times\!10^{-5}$  & $0.876$ & $23$      & $82.61$ \\
H11 & $5.21\!\times\!10^{-7}$  & $5.28\!\times\!10^{-7}$  & $1.014$ & $253$     & $90.12$ \\
H12 & $2.55\!\times\!10^{-6}$  & $2.99\!\times\!10^{-6}$  & --      & $0$       & --     \\
H13 & --                       & $5.36\!\times\!10^{-6}$  & --      & $1{,}213$ & $89.37$ \\
H14 & $1.99\!\times\!10^{-6}$  & $1.80\!\times\!10^{-6}$  & $0.905$ & $697$     & $90.10$ \\
H15 & $6.71\!\times\!10^{-7}$  & $6.77\!\times\!10^{-7}$  & $1.009$ & $800$     & $90.38$ \\
\midrule
Mean ($15$ heads) & $6.80\!\times\!10^{-6}$ & $6.59\!\times\!10^{-6}$ & $0.969$ & -- & $89.29 \pm 2.63$ \\
\bottomrule
\end{tabular}
\end{table}

The rank-2 reduction does not propagate to substitution: rank-2 perplexity exceeds rank-1 by $0.013$ nats. \gdn writes one rank-1 outer per step, so a rank-2 atom decomposes into rank-1 atoms the dictionary already covers, or compresses an $r$-step write history the cache cannot accept at firing level.

\subsection{Flat-SAE plus top-1 SVD substitution protocol}
\label{app:flat_svd}

The cross-architecture baseline trains a flat TopK SAE on $\mathrm{vec}(S_t)$ at the same cell as the architecture-matched \writesae, then reduces each firing's reconstructed state to its leading SVD outer product. Flat dimensions: \gdn $16{,}384$ ($128{\times}128$), Mamba-2-370M L24 H0 $8{,}192$ ($128{\times}64$), RWKV-7-1.5B L12 H0 matched to per-head outer write. Training matches \writesae{} ($n_\mathrm{feat}{=}2{,}048$, $k{=}32$, seed $42$, $20$ epochs, peak LR $3{\times}10^{-4}$). At evaluation we encode $S_t$, reshape into $\hat{S}_t$, take top-1 SVD outer $\hat{S}_t^{(1)} = \sigma_1 \mathbf{u}_1 \mathbf{v}_1^\top$, rescale to native Frobenius norm, and substitute. Per-firing replay caps firings per feature at three. Sample sizes: $n{=}1{,}851$ \gdn{}, $2{,}367$ Mamba-2, $6{,}591$ RWKV-7. Source JSONs at \texttt{flat\_sae\_svd\_\{gdn,mamba2,rwkv7\}/}.

\section{Alternative Explanations Considered}
\label{app:alternative-explanations}

\begin{figure}[!htbp]
  \centering
  \includegraphics[width=0.95\linewidth]{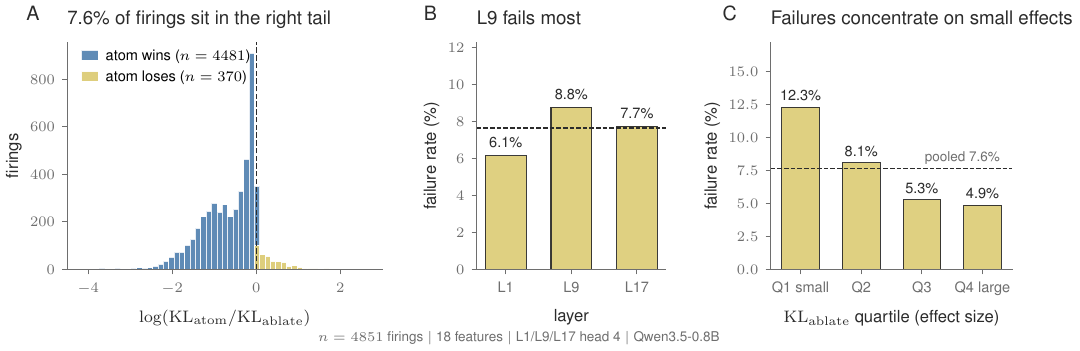}
  \caption{\textbf{Deletion beats the atom mainly on small-effect firings.} (a) $\log\mathrm{KL}_\mathrm{atom}/\mathrm{KL}_\mathrm{delete}$ over $n {=} 4{,}851$ firings (L1/L9/L17, 0.8B): $4{,}481$ atom wins, $370$ losses ($7.6\%$). (b) Per-layer failure rate close to the $7.6\%$ pooled mean. (c) Failure rate by $\mathrm{KL}_\mathrm{delete}$ effect-size quartile: Q1 $12.3\%$ to Q4 $4.9\%$.}
  \label{fig:failure_modes}
\end{figure}

Five readings could in principle reduce the partition to an artifact:
\begin{itemize}[topsep=2pt,itemsep=1pt,leftmargin=*]
\item \textbf{Decoder rank-1 prior}: App.~\ref{app:encoderswap} shows encoder-swap MSE within $8\%$ of the dense flat upper bound at L1/L9.
\item \textbf{Exemplar selection}: App.~\ref{app:cosine_free} cosine-free classifier replicates the main substitution result.
\item \textbf{Threshold fragility}: $92.4\%$ ordering holds within $\pm 2$pp across $0.5{\times}$--$2{\times}$ threshold sweeps, $10^4$-sample permutation $p<10^{-4}$, replicating at L1\,H4 + L17\,H4 on $n{=}4{,}851$ events.
\item \textbf{Small-model artifact}: Qwen3.5-4B L12 H8 reproduces the partition at $116\times$ register/null ratio, $27$B L32 H16 at $60\times$ (\Sref{sec:scope}, App.~\ref{app:scaling}).
\item \textbf{Random failure distribution}: Fig.~\ref{fig:failure_modes} shows the $7.6\%$ cases where deletion beats the atom concentrate on smallest-effect-size firings (Q1 $12.3\%$ vs Q4 $4.9\%$), consistent with the smallest effects being hardest to distinguish.
\end{itemize}

\subsection{Cosine threshold and mixture order}
\label{app:tau_k_sensitivity}

Sweeping $\tau$ and the GMM mixture order at L9 H4 does not change how often the atom beats deletion. The 0.8B count moves $1.10\times$ across $\tau \in \{0.02, 0.03, 0.05, 0.10\}$; at every $\tau$ the replacement win rate remains in the $90$--$94\%$ band. GMM $k{=}3$ adds a third component at weight $0.175$ but $\Delta\mathrm{BIC}(k{=}2 \to k{=}3){=}{-}4.15$ favors the two-component fit (bimodality coefficient $0.208$, threshold not reached). The $k$-sweep at L9 H4 runs $67/100/1{,}881$ register/bundle/null at $k{=}16$, $76/97/1{,}875$ at $k{=}32$, $155/137/1{,}756$ at $k{=}64$.

\begin{table}[h]
  \centering
  \small
  \caption{\textbf{Partition counts shift but replacement remains stable across $\tau$ and $k$.} Per-class win rate at L9 H4 is $91.25\%$ on the register pool chosen by replacement success and stays in $[90.7, 94.0]\%$ across the cosine partition's two classes.}
  \label{tab:tau_k_sensitivity}
  \setlength{\tabcolsep}{6pt}
  \begin{tabular}{lcccc}
    \toprule
    \emph{Cosine threshold $\tau$} & $0.02$ & $0.03$ & $0.05$ & $0.10$ \\
    \midrule
    Qwen3.5-0.8B L9 H4 register count & $243$ & $235$ & $222$ & $221$ \\
    Qwen3.5-0.8B L9 H4 bundle count   & $73$  & $81$  & $94$  & $95$ \\
    Qwen3.5-4B L12 H8 register count  & $227$ & $224$ & $220$ & $183$ \\
    Qwen3.5-4B L12 H8 bundle count    & $16$  & $19$  & $23$  & $60$ \\
    \midrule
    \emph{GMM mixture order $k$} & \multicolumn{2}{c}{$k{=}2$} & \multicolumn{2}{c}{$k{=}3$} \\
    \cmidrule(lr){2-3} \cmidrule(lr){4-5}
    BIC at L9 H4 & \multicolumn{2}{c}{$-679.18$ (preferred)} & \multicolumn{2}{c}{$-683.33$ ($\Delta{=}{-}4.15$)} \\
    \midrule
    \emph{$k$-sweep ($\nf{=}2{,}048$)} & $k{=}16$ & \multicolumn{2}{c}{$k{=}32$} & $k{=}64$ \\
    \midrule
    L9 H4 register / bundle / null & $67/100/1{,}881$ & \multicolumn{2}{c}{$76/97/1{,}875$} & $155/137/1{,}756$ \\
    \bottomrule
  \end{tabular}
\end{table}

\subsection{Cosine-free classifier reproduces the substitution result}
\label{app:cosine_free}

\paragraph{F758 demotion.}
F758 has the highest single-atom cosine ($0.985$) but fires at $5\times10^{-4}$ activation rate; its few hits came from an amplified scan at $5\times$ mean register activation. F1335 is the main-text exemplar because it fires on $5.24\%$ of validation tokens.

A replacement-success classifier on the L9 H4 feature pool that reads no decoder geometry (register if the atom beats deletion on more than half its firings, bundle otherwise) agrees with the cosine-classifier on 7 of 8 features that fired above threshold. The single disagreement is F87, the canonical bundle exemplar at $\widetilde{\cos}{=}0.012$, whose atom beats deletion on $94\%$ of its $300$ firings. The logit formula explains why two coordinates disagree at a single atom while agreeing at the class level.

The cosine-free pool reproduces the main result. The replacement-success register class fires $1{,}851$ times, with the atom beating deletion on $91.25\%$ of firings, the same number reported in \Sref{sec:mechvalid}. Cosine-register fires $1{,}551$ times at $90.72\%$; cosine-bundle is F87 alone at $94.00\%$. The two classes report the same per-firing rate within $3$pp; removing cosine from the pipeline leaves the $91\%$ result unchanged. The 8-feature pool was selected register-skewed (Cohen's $\kappa$ degenerates), so the relevant evidence is the per-class win rate. App.~\ref{app:population_kl} extends to $87$ atoms: $88.98\%$ on cosine-bundle ($n{=}94$), $91.36\%$ on cosine-register ($n{=}61$), Mann-Whitney $p{=}0.239$.

\begin{table}[h]
  \centering
  \small
  \caption{\textbf{Class-level claims replicate across SAE seeds; per-atom identity does not.}}
  \label{tab:claims}
  \setlength{\tabcolsep}{4pt}
  \begin{tabular}{@{}p{0.32\linewidth}p{0.14\linewidth}p{0.50\linewidth}@{}}
    \toprule
    Claim & Granularity & Evidence \\
    \midrule
    Register/bundle partition & Class-level & GMM $\Delta\mathrm{BIC}{=}-296$; CV $4$--$12\%$ across 3 seeds \\
    Partition scale-invariance & Class-level & $34\times$ Qwen range; DeltaNet sweep \\
    Register firings direction-selective & Class-level & Selectivity $0.9953$ across $47/48$ cells \\
    Atom beats deleting the write & Firing-level & $92.4\%$ of $n{=}4{,}851$; permutation $p<10^{-4}$, Cliff's $\delta = +0.825$. Population on $87$ atoms: $89.80\%$, CI $[88.1, 91.3]$ \\
    Individual atom identity (e.g., F758) & Per-atom & $<1\%$ basis overlap across seeds (illustrative) \\
    \bottomrule
  \end{tabular}
\end{table}

\subsection{DeltaNet substitution failure: geometric and causal decoupling}
\label{app:deltanet_decoupling}

DeltaNet L12 H8 produces the largest register-cosine separation (register median $\cos{=}0.997$, register/null $383\times$ at $k{=}128$; Tab.~\ref{tab:cross_arch}), yet the firing-level substitution test on the same cell fails. On $n{=}395$ firings across $37$ passages, the strict ordering $\mathrm{KL}_\mathrm{atom}<\mathrm{KL}_\mathrm{delete}<\mathrm{KL}_\mathrm{random}$ holds on $17.7\%$ of firings (CI $[14.0, 21.5]$), against $89.5\%$ on the Qwen3.5-0.8B L9 H4 \gdn cell. The atom beats deletion on $48.3\%$ of firings, indistinguishable from chance. Median $\mathrm{KL}_\mathrm{atom}{=}2.70\!\times\!10^{-4}$ exceeds median $\mathrm{KL}_\mathrm{delete}{=}2.30\!\times\!10^{-4}$.

DeltaNet runs the bilinear write rule with the convex gate off (\texttt{use\_gate=false}), so $S_t$ accumulates without per-position decay. Top-1 singular variance is $97.79\%$ (stable rank $1.023$), so the state is geometrically rank-1, but the dominant singular direction integrates every prior write. The atom matches the average direction of $\mathbf{k}_t \mathbf{v}_t^\top$ across positions; under no decay, the cache at firing time is set by integrated history, not the local write the atom is trained to recover. Adding a gate (Qwen3.5 \gdn) restores per-position decay and the replacement test recovers (Tab.~\ref{tab:cross_arch}, atom beats deletion on $> 89\%$ throughout the $34\times$ Qwen range).

\section{Register vs bundle: systematic differences beyond population win rate}
\label{app:register_bundle_systematic}

The population test in App.~\ref{app:population_kl} reports a null gap in how often the atom beats deleting the write ($91.4\%$ vs $89.0\%$, $p{=}0.24$). Beyond binary win/loss, the two classes differ on quantities that affect interpretive use of the partition. We re-use the $87$ alive atoms from the L9 H4 population KL run; $23$ register and $56$ bundle atoms enter the top-$1$ analyses, with $2{,}190$ records carrying per-firing top-$1$ token id, next-token rank, and log-probability under each cache state.

\begin{table}[!htbp]
\centering
\small
\caption{\textbf{Register and bundle classes differ on three of four causal axes at L9 H4.} Top-$1$ disrupt is the fraction of firings on which the patched cache flips the model's top-$1$ next-token. $\Delta$KL is per-atom median KL\textsubscript{delete} $-$ KL\textsubscript{atom} (nats). Firings per atom is the activation count over $10{,}000$ validation tokens. Cliff's $\delta$ positive: register $>$ bundle. Mann-Whitney $p$ two-sided.}
\label{tab:register_bundle_systematic}
\begin{tabular}{lcccc}
\toprule
Axis & Register median & Bundle median & Cliff's $\delta$ & MW $p$ \\
\midrule
$\Delta$KL (nats) & $0.682$ & $0.579$ & $+0.109$ & $0.45$ \\
Top-$1$ disrupt under deletion & $0.700$ & $0.633$ & $+0.294$ & $\mathbf{0.041}$ \\
Top-$1$ disrupt under atom & $0.423$ & $0.400$ & $+0.161$ & $0.26$ \\
Firings per atom (population scan) & $100$ & $70.5$ & $+0.518$ & $\mathbf{<10^{-3}}$ \\
\bottomrule
\end{tabular}
\end{table}

\paragraph{Top-$1$ disruption.}
Removing a register flips the top-$1$ on $70.0\%$ of firings, against $63.3\%$ for a bundle (Cliff's $\delta{=}+0.29$, $p{=}0.041$); cosine vs deletion top-$1$ flip-rate Pearson $r{=}+0.25$ ($p{=}0.027$). The atom-state top-$1$ disrupt is statistically tied across classes, so the rank-1 atom recovers the native top-$1$ at comparable rates whether the underlying write is cosine-aligned or not.

\paragraph{Firing breadth.}
Registers fire on a median of $100$ validation tokens; bundles on $70.5$ ($\delta{=}+0.52$, $p{<}10^{-3}$). Cosine-vs-firing-count Spearman $\rho{=}+0.73$ ($p{<}10^{-5}$) is the strongest continuous correlation in this analysis. The interpretive consequence: register atoms supply more firing examples per unit of validation traffic, making them cheaper to recruit for circuit-tracing or edit experiments.

\paragraph{$\Delta$KL effect size.}
Median $\Delta$KL is $0.682$ nats for registers and $0.579$ for bundles (Cliff's $\delta{=}+0.11$, $p{=}0.45$). The two classes inflict comparable lesion magnitude, consistent with the null on win rate.

\paragraph{Behavioral wedge beyond F87.}
The cosine-free flip of F87 (App.~\ref{app:cosine_free}, atom beats deletion on $94\%$ of firings) is one point on the continuous gradient above. The cosine partition predicts which atoms carry more next-token mass and how often they fire, even though it does not predict whether a given firing's substitution beats deleting the write; the firing-level test holds across the full alive dictionary at $89.8\%$.

\section{Extended Results: Selectivity, Reader Traces, and Residual-SAE Comparison}
\label{app:extended_results}

\subsection{Selectivity sanity check and orthogonal-rank-1 control}
\label{app:selectivity_sanity}

An orthogonal rank-1 control at matched Frobenius norm returns selectivity $0.998$; any matched-norm rank-1 perturbation, aligned or orthogonal, scores above $0.99$, so top-$K$-overlap selectivity cannot separate the SAE atom from a random rank-1 (Fig.~\ref{fig:selectivity_law}). Top-$K$-overlap selectivity averages $\mathbf{0.9953}$ across $47/48$ cells under matched-norm random rank-1 perturbation (CI $[0.9930, 0.9976]$). The distinguishing evidence is firing-level KL ordering, the population substitution test (App.~\ref{app:population_kl}), and the amplitude-conditional F87 inversion.

\subsection{Reader traces at L9 H4 exemplars}
\label{app:reader_traces}

Register atoms read into specific later attention heads at $3\text{--}7\times$ baseline (F53 into L5 H5 at $6.9\times$, F63 into L17 H4 at $5.2\times$, F1335 into L21 H10 at $7.5\times$); the signal does not diffuse across the residual stream. These three exemplars show reader-enriched pathways, but we do not claim generality.

\subsection{Residual-stream SAE on the same model}
\label{app:residual_sae_comparison}

A residual-stream SAE asks a different question: its cosine measures alignment between an atom and the activation the dictionary was trained to reconstruct, an alignment the TopK objective guarantees \citep{gao2024scaling, sun2025densefeat}. On Qwen3.5-0.8B, a TopK SAE on the L15 attention output ($n_\mathrm{feat}{=}2{,}048$, $k{=}32$) returns $1{,}848$ register atoms against $27$ bundles at register median $\cos{=}0.21$ ($\Delta\mathrm{BIC}{=}+306.5$), against $222$ and $94$ on the GDN cache. The substitution test instead measures cosine to $\mathbf{k}_t\mathbf{v}_t^\top$, an object the dictionary was not trained against. Residual-stream atoms are vectors; they cannot occupy the cache slot the rank-1 atom replaces.

\subsection{Memory-edit intervention}
\label{app:memory_edit}

The cache edit reported in \Sref{sec:steering} runs at L9 H4 on the same dictionary used for the firing-level KL ordering. Records are at \hfrepolink{results/memory\_edit\_F412/} as \texttt{summary.json}, \texttt{records.jsonl}, \texttt{dose\_response.csv}, \texttt{feature\_detection.json}.

\paragraph{Feature detection.}
For each F412 firing we sum $\log p_\mathrm{native} - \log p_\mathrm{delete}$ across firings per token id; the highest-scoring token is the feature's preferred token. F412's winner is Qwen tokenizer id $98818$ (gloss ``space''), summed native-minus-delete difference $14.5$ nats. Median natural activation $a^\ast{=}0.31$.

\paragraph{Erasure at natural firings.}
At $n{=}150$ natural firing positions, deleting the write changes the preferred token's log probability by median $-0.116$ nats, $95\%$ CI $[-0.265, -0.042]$, paired Wilcoxon $p{=}1.07\times 10^{-6}$, mean $-0.349$. Preferred-token rank shifts from native median $68{,}485$ to patched median $77{,}444$.

\begin{table}[!htbp]
\centering
\small
\caption{\textbf{Erasure at $n{=}150$ natural firings of F412 (Qwen3.5-0.8B L9 H4) reduces the atom's preferred next-token probability.} Preferred token: Qwen id $98818$ (``space'').}
\label{tab:memory_edit_erase}
\begin{tabular}{lc}
\toprule
Quantity & Value \\
\midrule
$n$ firings & $150$ \\
Median $\Delta\log p$ (preferred token) & $-0.116$ nats \\
$95\%$ CI & $[-0.265, -0.042]$ \\
Mean $\Delta\log p$ & $-0.349$ nats \\
Paired Wilcoxon $p$ & $1.07\times 10^{-6}$ \\
Native median rank (preferred token) & $68{,}485$ \\
Patched median rank (preferred token) & $77{,}444$ \\
\bottomrule
\end{tabular}
\end{table}

\paragraph{Cache edit at non-firing positions.}
At $n{=}150$ non-firing positions we add $m\cdot a^\ast \cdot \mathbf{v}_{F412}\mathbf{w}_{F412}^\top$ to the cache. No tested magnitude reaches significance; per-position writes at non-firing slots are dominated by surrounding context. Dose-response over $n{=}50 \times 6$ magnitudes returns a weakly monotone trend with no significant linear fit.

\begin{table}[!htbp]
\centering
\small
\caption{\textbf{Cache edits at $n{=}150$ non-firing positions do not reach significance.}}
\label{tab:memory_edit_install}
\begin{tabular}{lcc}
\toprule
Magnitude & Median $\Delta\log p$ & $p$ \\
\midrule
$1\times a^\ast$ & $+3.1\times 10^{-5}$ & $0.67$ \\
$2\times a^\ast$ & $+0.008$ & $0.28$ \\
$4\times a^\ast$ & $+0.016$ & $0.15$ \\
\bottomrule
\end{tabular}
\end{table}

\subsection{Single-position sign test}
\label{app:predictive_steering}

For each target token $T$ we compute $v_T^\ast = W_O[\mathrm{head}]^\top W_U[T]/\|\cdot\|$, write that direction at one cache position, and measure the target token's logit change. Pooled run $n{=}2{,}000$ triples at L9 H4: pooled $R^2 = -0.06$ (measured scale varies with prompt context), directional agreement $84.6\%$ CI $[83.0,86.2]$, Pearson $r{=}0.162$ ($p{=}3.7\times10^{-13}$), median measured/predicted ratio $1.08$.

\subsection{Generation intervention: full results}
\label{app:behavioral_steering}

The direction $v_{T}^{\ast} = W_O[\text{head}]^\top W_U[T]/\|\cdot\|$ is chosen to increase the logit of target token $T$ under a rank-1 cache write. Sweep: $30$ prompts $\times$ $5$ cache positions $\times$ $8$ targets $\times$ $3$ magnitudes $= 3{,}600$ trials at L9 H4. Each edit is applied across three consecutive cache positions at $m \cdot \|\mathbf{k}_t \mathbf{v}_t^\top\|$; the model then generates $20$ tokens greedily. Targets are stratified by their initial rank under the unmodified model: \emph{frequent} (${\sim}17{,}000$), \emph{rank 100--1000}, \emph{rare} ($\geq 10{,}000$), and \emph{semantic} (out-of-context).

\begin{table}[!htbp]
\centering
\small
\caption{\textbf{Pooled target-appearance lift by magnitude across $n{=}1{,}200$ trials per row.} Lift subtracts the unmodified model's rate ($8.3\%$). $3{\times}$ gives the largest measured lift.}
\label{tab:behavioral_steering_magnitude}
\begin{tabular}{lccccc}
\toprule
Magnitude & target appears after edit & lift (pp) & rank improved & median logp lift (nats) \\
\midrule
$1.5\times$ & $16.7\%$ & $+8.3$pp & --- & --- \\
$3.0\times$ & $\mathbf{25.0\%}$ & $\mathbf{+16.7}$pp & $\mathbf{77.4\%}$ & $\mathbf{+1.27}$ \\
$6.0\times$ & $16.7\%$ & $+8.3$pp & --- & --- \\
\bottomrule
\end{tabular}
\end{table}

\begin{table}[!htbp]
\centering
\small
\caption{\textbf{Breakdown by target class at $m{=}3{\times}$ ($n{=}300$ per class).} Targets initially ranked $100$--$1000$ reach $100\%$ vs $33.3\%$ under the unmodified model ($+66.7$pp). Out-of-context classes show $4{,}039$--$17{,}526$-position rank shifts but $0\%$ target appearance because greedy generation cannot promote a starting rank ${\geq} 17{,}000$ to top-$1$ over three positions.}
\label{tab:behavioral_steering_class}
\begin{tabular}{lccccc}
\toprule
Class & initial rank & edited & native & lift (pp) & median rank shift \\
\midrule
frequent (out-of-context) & ${\sim}17{,}000$ & $0\%$ & $0\%$ & $0$ & $17{,}526$ \\
\textbf{rank 100--1000} & $100\text{--}1000$ & $\mathbf{100\%}$ & $33.3\%$ & $\mathbf{+66.7}$ & $517$ \\
rare & $\geq 10{,}000$ & $0\%$ & $0\%$ & $0$ & $16{,}800$ \\
semantic (out-of-context) & --- & $0\%$ & $0\%$ & $0$ & $4{,}039$ \\
\bottomrule
\end{tabular}
\end{table}

\paragraph{Rank 100--1000 result.}
For targets initially ranked $100$--$1000$ under greedy decoding, the $m{=}3{\times}$ cache edit yields $300/300$ continuations with the targeted token. Pooled across all four classes the lift is $+16.7$pp ($25.0\%$ vs $8.3\%$ native, $n{=}1{,}200$); restricted to the rank-100-to-1000 class, the lift is $+66.7$pp.

\paragraph{Stratification note.}
The frequent, rare, and semantic classes were drawn from out-of-context vocabulary items that the unmodified model ranks at least $17{,}000$. They produce $0\%$ target appearance at every magnitude. The direction chosen by the formula shifts these tokens by $4{,}039$--$17{,}526$ rank positions; the underlying logit signal appears, but greedy decoding over three cache positions cannot move the starting rank to top-$1$. The rank-100-to-1000 class meets both conditions; out-of-context classes meet only the first.

\paragraph{Magnitude saturation.}
Pooled lift is non-monotone: $+8.3$pp at $1.5\times$, $+16.7$pp at $3.0\times$, $+8.3$pp at $6.0\times$. Beyond the best measured magnitude, the cache write dominates surrounding context and degrades the rest of generation, mirroring the newline-rate saturation at $10\times$ in \Sref{sec:steering}.

\section{Reproducibility}
\label{app:repro}

\paragraph{Code, checkpoints, license.}
All scripts that produce the reported numbers, tables, and figures are in the repo snapshot at \repolink. Trained SAE checkpoints, cached \gdn{} state tensors, and per-head deletion-control JSON outputs are on HuggingFace at \hfreporootlink (four SAE variants $\times$ Qwen3.5-0.8B/4B/27B; 0.8B covers L9 and L1\,H4, L17\,H4). Code and checkpoints under MIT; base models under Tongyi Qianwen. $\sim\!180$ H100-hours single-GPU; one canonical SAE config fits in $\sim\!6$ H100-hours. Reference container \texttt{pytorch/pytorch:2.4.1-cuda12.1-cudnn9-runtime} with pinned \texttt{transformers}, \texttt{flash-linear-attention}~\citep{yang2024fla}, \texttt{datasets}, \texttt{h5py}, \texttt{huggingface\_hub}; \texttt{pip install -e .} reproduces outside the container.

\paragraph{Datasets.}
\label{app:datasets}
OpenWebText \citep{Gokaslan2019OpenWeb} streamed from \texttt{Skylion007/openwebtext} (CC0). Tokenize with the Qwen3.5 tokenizer (152K BPE, shared across 0.8B/4B/27B), pack into $1{,}024$-token blocks. \gdn{} on $5{,}000\times 1{,}024$ sequences yields $\approx\!5\!\times\!10^{6}$ matrix-valued samples per head, $80/20$ train/val at seed $42$. Evaluation pulls disjoint shards: $500$ sequences for the cache-replacement PPL sweep (positions $0$--$511$ context, teacher-forced loss on $512$--$1023$) and $20$ paired-connector passages for the bits/token comparison (Fig.~\ref{fig:causal_dashboard}b). The $4$B generation probe uses a third $40$-prompt set, disjoint from training, PPL, and connector splits.

\paragraph{Per-firing atom selection.}
\label{app:atom_select}
The firing-level substitution test in \Sref{sec:mechvalid} uses the dominant atom of the SAE's TopK encoding at each firing position:
\begin{verbatim}
for each firing position t of feature i:
    a   = SAE.encode(S_t)                          # k = 32 nonzeros
    j   = argmax_{r in TopK(a)} a_r                # dominant atom
    A_j = decoder_v[j] @ decoder_w[j].T            # rank-1 outer
    sigma_t = norm(beta_t * k_t @ v_t.T, 'fro') / norm(A_j, 'fro')
    write_substitute = sigma_t * A_j               # matched norm
    forward(state := previous + write_substitute)  # atom/delete/random
\end{verbatim}
The deletion condition writes a zero outer product at matched Frobenius norm; the random rank-1 condition draws from the matched-norm random routine. \gdn{} mutates its cache in place, so per-firing replay deep-copies the cache for each conditional forward pass.

\paragraph{Implementation notes.}
\label{app:impl_notes}
Alive-feature counts on the dense-encoder \writesae{} stay in the $300$--$1{,}200$ range only with both regularizers: an auxiliary dead-feature loss ($\lambda_\mathrm{aux}{=}10^{-2}$) reconstructs residuals through atoms silent for $100$ steps, and a resampler fires every $250$ steps for atoms still silent after the auxiliary term. Dropping either roughly doubles the dead-atom rate. The dense encoder reaches the lowest validation MSE on every cell tested but its atoms do not carry a clean rank-1 read; the bilinear matched-filter encoder $a_i = \mathbf{v}_i^\top \state\,\mathbf{w}_i$ trails MSE by $5$--$15\%$ but its firing coefficient matches the state's projection onto the same rank-1 direction the decoder writes, so we use it for the 4B generation probe. Skipping the cache deep-copy biases results toward whichever condition runs last (cost $\sim 1.4\times$ wall-clock per firing).

\end{document}